\documentclass[sigconf,authorversion,nonacm]{acmart}

\usepackage{graphicx}
\usepackage{amsmath}
\usepackage{booktabs}
\usepackage{multirow}
\usepackage[inline]{enumitem}
\usepackage{multicol}
\usepackage{subcaption}
\usepackage{hyperref}
\usepackage[capitalize]{cleveref}
\Crefname{section}{Section}{Sections}
\Crefname{table}{Table}{Tables}
\crefname{figure}{Figure}{figures}
\newcommand{\phylonn}{\emph{Phylo-NN}}

\newcommand{\z}{\mathbf{z}}
\newcommand{\zq}{\mathbf{z}^Q}

\newcommand{\imagein}{\mathbf{x}}
\newcommand{\imageout}{\mathbf{\hat{x}}}

\newcommand{\chin}{\mathit{C}} 
\newcommand{\win}{\mathit{W}} 
\newcommand{\hin}{\mathit{H}} 

\newcommand{\chphylo}{\mathit{C_{\text{p}}}}
\newcommand{\zphylo}{\mathit{\mathbf{z}_{\text{p}}}}

\newcommand{\znonattr}{\mathit{\mathbf{z}_{\text{np}}}}

\newcommand{\zphyloq}{\mathit{\mathbf{z}_{\text{p}}^Q}}

\newcommand{\encoder}{\mathit{E}}
\newcommand{\decoder}{\mathit{D}}
\newcommand{\transformer}{\mathit{T}}

\newcommand{\phyloencoder}{\mathit{PE}}
\newcommand{\phylodecoder}{\mathit{PD}}

\newcommand{\znonattrq}{\mathit{\mathbf{z}_{\text{np}}^Q}}

\newcommand{\embeddingsize}{\mathit{d}}
\newcommand{\nlevelsphylo}{\mathit{n_\text{l}}}
\newcommand{\ncodebook}{\mathit{n_\text{q}}}
\newcommand{\codesperlevel}{\mathit{n_\text{p}}}
\newcommand{\nonattrcodes}{\mathit{n_{\text{np}}}}

\newcommand{\Hphylo}{\mathit{H_{\text{p}}}}
\newcommand{\Hnonattr}{\mathit{H_{\text{\text{np}}}}}

\newcommand{\CE}{\mathbf{\text{CE}}}

\newcommand{\orthogconv}{\mathit{L_{\text{o}}}}

\newcommand{\advloss}{\mathit{L_{\text{adv}}}}
\newcommand{\advmlp}{\mathit{MLP_{\text{adv}}}}

\newcommand{\lquant}{\mathit{L_{\text{q}}}}
\newcommand{\lrec}{\mathit{L_{\text{rec}}}}
\newcommand{\lphylo}{\mathit{L_{\text{p}}}}

\newcommand{\zconceptwhitening}{\mathit{z_{\text{cw}}}}

\AtBeginDocument{%
  \providecommand\BibTeX{{%
    \normalfont B\kern-0.5em{\scshape i\kern-0.25em b}\kern-0.8em\TeX}}}

\begin{document}

\title{Discovering Novel Biological Traits From Images Using Phylogeny-Guided Neural Networks}

\author{Mohannad Elhamod} \orcid{0000-0002-2383-947X}
\affiliation{%
  \institution{Virginia Tech}
  \country{}
}
\email{elhamod@vt.edu}

\author{Mridul Khurana} \orcid{0009-0003-9346-3206}
\affiliation{%
  \institution{Virginia Tech}
  \country{}
}
\email{mridul@vt.edu}

\author{Harish Babu Manogaran} \orcid{ 0000-0003-3709-4656}
\affiliation{%
  \institution{Virginia Tech}
  \country{}
}
\email{harishbabu@vt.edu}

\author{Josef C. Uyeda} \orcid{0000-0003-4624-9680}
\affiliation{%
  \institution{Virginia Tech}
  \country{}
}
\email{juyeda@vt.edu}

\author{Meghan A. Balk} \orcid{0000-0003-2699-3066}
\affiliation{%
  \institution{Battelle}
  \country{}
}
\email{meghan.balk@gmail.com}

\author{Wasila Dahdul} \orcid{0000-0003-3162-7490}
\affiliation{%
  \institution{University of California, Irvine}
  \country{}
}
\email{wdahdul@uci.edu}

\author{Yasin Bakış} \orcid{0000-0001-6144-9440}
\affiliation{%
  \institution{Tulane University}
  \country{}
}
\email{ybakis@tulane.edu}

    \author{Henry L. Bart Jr.} \orcid{0000-0002-5662-9444}
\affiliation{%
  \institution{Tulane University}
  \country{}
}
\email{hbartjr@tulane.edu}

\author{Paula M. Mabee} \orcid{0000-0002-8455-3213}
\affiliation{%
  \institution{Battelle}
  \country{}
}
\email{mabee@battelleecology.org}

\author{Hilmar Lapp} \orcid{0000-0001-9107-0714}
\affiliation{%
  \institution{Duke University}
  \country{}
}
\email{Hilmar.Lapp@duke.edu}

\author{James P. Balhoff}  \orcid{0000-0002-8688-6599}
\affiliation{%
  \institution{University of North Carolina at Chapel Hill}
  \country{}
}
\email{balhoff@renci.org}

\author{Caleb Charpentier} \orcid{0000-0002-9787-7081}
\affiliation{%
  \institution{Virginia Tech}
  \country{}
}
\email{calebc22@vt.edu}

\author{David Carlyn} \orcid{0000-0002-8323-0359}
\affiliation{%
  \institution{The Ohio State University}
  \country{}
}
\email{carlyn.1@osu.edu}

\author{Wei-Lun Chao} \orcid{0000-0003-1269-7231}
\affiliation{%
  \institution{The Ohio State University}
  \country{}
}
\email{chao.209@osu.edu}

\author{Charles V. Stewart} \orcid{0000-0001-6532-6675}
\affiliation{%
  \institution{Rensselaer Polytechnic Institute}
  \country{}
}
\email{stewart@rpi.edu}

\author{Daniel I. Rubenstein}\orcid{0000-0001-9049-5219}
\affiliation{%
    \institution{Princeton University}
  \country{}
}
\email{dir@princeton.edu}

\author{Tanya Berger-Wolf} \orcid{0000-0001-7610-1412}
\affiliation{%
  \institution{The Ohio State University}
  \country{}
}
\email{berger-wolf.1@osu.edu}

\author{Anuj Karpatne} \orcid{0000-0003-1647-3534}
\affiliation{%
  \institution{Virginia Tech}
  \country{}
}
\email{karpatne@vt.edu}




\renewcommand{\shortauthors}{Elhamod et al.}

\begin{abstract}
Discovering evolutionary traits that are heritable across species on the tree of life (also referred to as a phylogenetic tree) is of great interest to biologists to understand how organisms diversify and evolve. However, the measurement of traits is often a subjective and labor-intensive process, making \textit{trait discovery} a highly label-scarce problem. We present a novel approach for discovering evolutionary traits directly from images without relying on trait labels. Our proposed approach, \phylonn{}, encodes the image of an organism into a sequence of quantized feature vectors--or codes--where different segments of the sequence capture evolutionary signals at varying ancestry levels in the phylogeny. We demonstrate the effectiveness of our approach in producing biologically meaningful results in a number of downstream tasks including species image generation and species-to-species image translation, using fish species as a target example.\footnote{The code and datasets for running all the analyses reported in this paper can be found at \url{https://github.com/elhamod/phylonn}.}
\end{abstract}

\ccsdesc[300]{Computing methodologies~Neural networks}
\ccsdesc[500]{Applied computing~Imaging}
\ccsdesc[500]{Computing methodologies~Computer vision}
\ccsdesc[500]{Computing methodologies~Image representations}
\ccsdesc[500]{Computing methodologies~Neural networks}
\ccsdesc[500]{Applied computing~Imaging}
\ccsdesc[300]{Computing methodologies~Regularization}


\keywords{computer vision, neural networks, phylogeny, morphology, knowledge-guided machine learning}




\maketitle

\section{Introduction}
\label{sec:intro}

One of the grand challenges in biology is to find features of organisms--or \textit{traits}--that define groups of organisms, their genetic and developmental underpinnings, and their interactions with environmental selection pressures \cite{houlerossoni2022}. Traits can be physiological, morphological, and/or behavioral (e.g., {beak color, stripe pattern, and fin curvature}) and are integrated products of genes and the environment. The analysis of traits is critical for predicting the effects of environmental change or genetic manipulation, and to understand the process of evolution. For example, discovering traits that are heritable across species on the tree of life (also referred to as the \textit{phylogenetic tree}), can serve as a starting point for linking traits to underlying genetic factors. Traits with such genetic or phylogenetic signal, termed \textit{evolutionary traits}, are of great interest to biologists, as the history of genetic ancestry captured by such traits can guide our understanding of how organisms diversify and evolve. This understanding enables tasks such as estimating the morphological features of ancestors, understanding how species have responded to environmental changes, and even predicting the potential future course of trait changes \cite{lynch1991, collyer2021}. However, the measurement of traits is not straightforward and often relies on subjective and labor-intensive human expertise and definitions~\cite{simoes2017giant}. Hence, \textit{trait discovery} has remained a highly label-scarce problem, hindering rapid scientific advancement \cite{lurig2021}.

With the recent availability of large-scale image repositories containing {millions of images} of biological specimens \cite{van2018inaturalist,singer2018survey,wah2011caltech}, there is a great opportunity for machine learning (ML) to contribute to the problem of trait discovery \cite{lurig2021}. In particular, advances in deep learning have enabled us to extract useful information from images and to map them to structured feature spaces where they can be manipulated in a number of ways. We ask the question: \textit{how can we develop deep learning models to discover novel evolutionary traits automatically from images without using trait labels?}

\begin{figure}[t]
  \centering
  \includegraphics[width=1.0\linewidth]{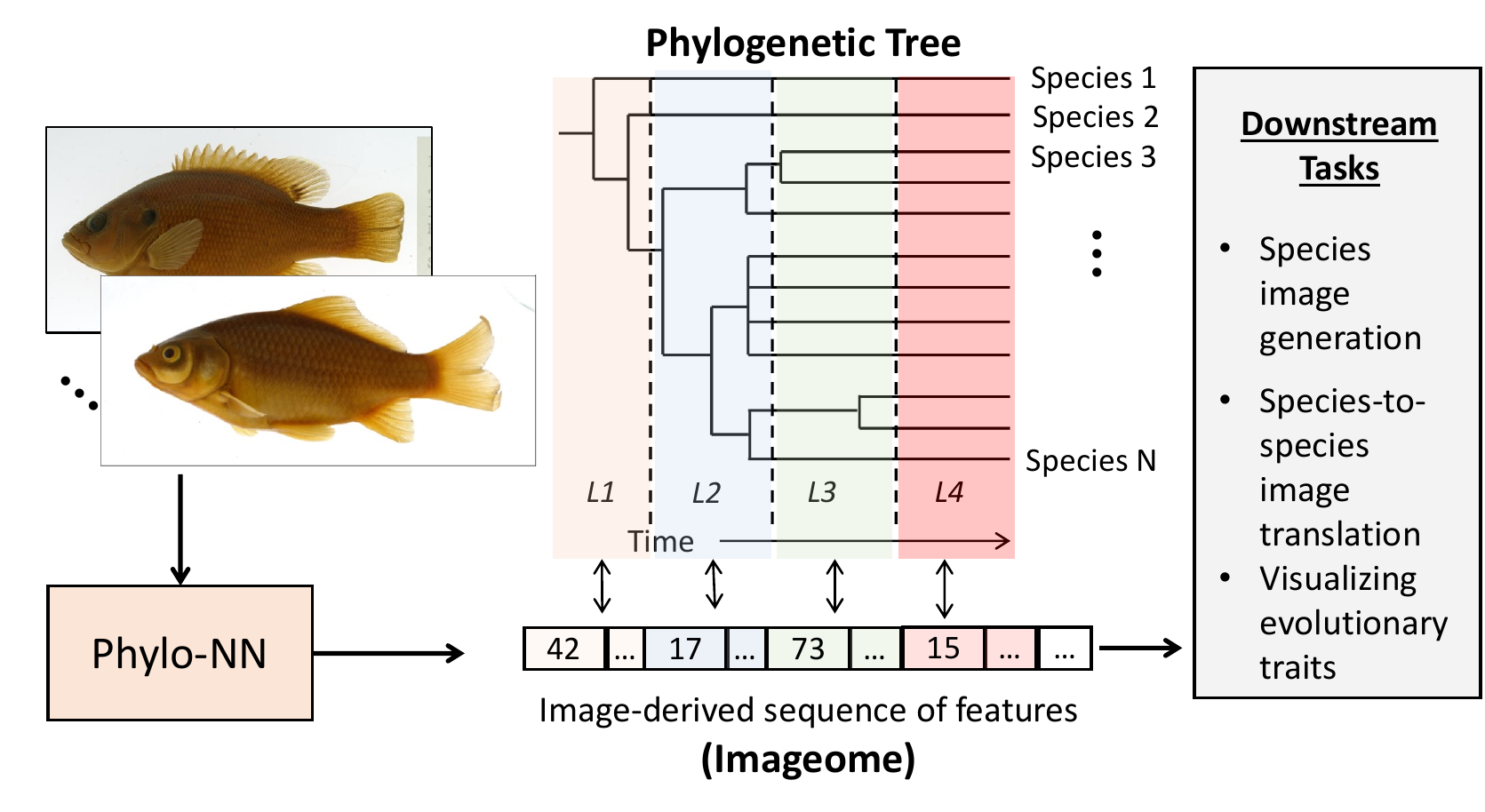}
  \caption{\phylonn{} converts images to discrete sequences of features (called Imageomes) where different sequence segments (shown in distinct colors) capture evolutionary information at varying ancestry levels of phylogeny (L1 to L4).}
  \label{fig:toc}
  \vspace{-4ex}
\end{figure}

Despite the biological relevance of this question, answering it is challenging for two main reasons. First, not all image features extracted by a deep learning model for ML tasks 
such as image reconstruction or species classification will exhibit evolutionary signals. Hence, it is important to disentangle the image features of an organism that preserve evolutionary information, from remaining features influenced by unrelated factors \cite{collyer2021}. Second, information about evolutionary signals is not available as a set of known attributes (or trait labels) but rather in the form of structured knowledge of how species are related to each other in the phylogenetic tree (see \cref{fig:toc}). Without access to trait labels, current methods for feature disentanglement in deep learning \cite{li2020latent,chen2020concept} are unfit for discovering evolutionary traits. Furthermore, current standards in deep learning for generative modeling \cite{richardson2021encoding,esser2021taming} or interpretable ML \cite{protopnet,nauta2021neural} are unable to leverage structured forms of biological knowledge (e.g., phylogenetic trees) in the learning of image features, and hence are unable to analyze and manipulate learned features in biologically meaningful ways.


We propose a novel approach for discovering evolutionary traits automatically from images, termed \textit{phylogeny-guided neural networks} (\phylonn{}), which encodes the image of an organism into a sequence of quantized feature vectors or ``codes'' (see \cref{fig:toc}). A unique feature of the image-derived sequences learned by \phylonn{} is that different segments of the sequence capture evolutionary information at varying ancestry levels in the phylogeny, where every level corresponds to a certain point of time in evolutionary history. 
Analogous to how the genome of an organism encodes all its genetic information and structures it as a set of genes, our image-derived sequences encodes all of the visual information contained in the organism's image and structures it as a set of evolutionary traits shared with ancestor nodes within its lineage (at different levels of phylogeny). 
We thus refer to the image-derived sequences of \phylonn{} as \textit{Imageomes}, a brand-new concept in evolutionary biology. By analyzing and manipulating the codes in the Imageomes of organisms, we can perform a number of biologically meaningful downstream tasks such as species image generation, species-to-species image translation, and visualization of evolutionary traits. We demonstrate the effectiveness of \phylonn{} in solving these tasks using fish species as a target example. 

Our work, for the first time, provides a bridge between the ``language of evolution'' represented as phylogeny and the ``language of images'' extracted by \phylonn{} as Imageomes. This work is part of a larger-scale effort to establish a new field of research in ``Imageomics'' \cite{imageomics}, where images are used as the source of information to accelerate biological understanding of traits, ranging from their selective consequences to their causation. Our work also provides a novel methodological advance in the emerging field of knowledge-guided machine learning (KGML) \cite{karpatne2022knowledge,karniadakis2021physics,karpatne2017theory} by using tree-based knowledge to structure the embedding space of neural networks and produce scientifically meaningful image generation and translation results.

\section{Background and Related Work}
\noindent \textbf{What is a Phylogenetic Tree?}
\label{sec:phylogeny}
A phylogenetic tree visually characterizes the evolutionary distances among a set of species and their common ancestors represented as nodes of the tree. In this tree, the length of every edge is a value that represents the evolutionary distance between two nodes (measured in time intervals representing thousands or millions of years), which is estimated from living species and time-calibrated ages using dated fossil ancestors or molecular methods. While rates of change along different edges may vary substantially, on average we expect that longer edges will accumulate higher levels of evolutionary trait change than shorter edges. In our work, we consider discretized versions of the phylogenetic tree with $\nlevelsphylo{}=4$ ancestry levels, such that every species class (leaf node in the tree) has exactly $\nlevelsphylo{} - 1$ ancestors. Every ancestry level corresponds to a certain point of time in evolutionary history. See \cref{app:phylo} for details on phylogeny preprocessing.

\noindent \textbf{Generative Modeling for Images:}
There exists a large body of work in deep learning for image generation, including methods based on Variational Autoencoders (VAEs) \cite{kingma2013auto}, Generative Adversarial Networks (GANs) \cite{goodfellow2020generative,stylegan,stylegan2,stylegan3,richardson2021encoding}, Transformer networks \cite{dosovitskiy2020image,he2022masked}, and Diffusion models \cite{croitoru2023diffusion}. 
While some recent advances in this field (e.g., DALL-E 2) have been shown to produce images with very high visual quality, they involve large and complex embedding spaces that are difficult to structure and analyze using tree-based knowledge (e.g., phylogeny). Instead, we build upon a recent line of work in generative modeling using vector-quantized feature representations of images \cite{vqvae, esser2021taming} that are easier to manipulate than continuous features. 
In particular, a recent variant of the VAE, termed Vector-Quantized VAE (VQVAE) \cite{vqvae}, uses discrete feature spaces quantized using a learned codebook of feature vectors and employs a PixelCNN \cite{van2016conditional} model for sampling in the discrete feature space. 
This work was extended in \cite{esser2021taming} to produce VQGAN, which is different from VQVAE in two aspects. First, it adds a discriminator to its framework to improve the quality of the generated images. Second, it uses a Transformer model, namely the GPT architecture \cite{radford2019language}, to generate images from the quantized feature space instead of a PixelCNN. VQGAN is a state-of-the-art method that generates images of better quality efficiently at higher resolutions than other counterparts such as StyleGAN \cite{stylegan3} and Vision Transformers \cite{dosovitskiy2020image,he2022masked}. Our work draws inspiration from VQGAN to embed images in discrete feature spaces (analogous to the discrete nature of symbols used in genome sequences) but with the grounding of biological knowledge available as phylogenetic trees.


\noindent \textbf{Interpretable ML:}
There is a growing trend in the ML community to focus on the interpretability of deep learning features \cite{du2019techniques}. Some of the earliest works in this direction include the use of saliency scores \cite{simonyan2013deep} and Class Activation Maps (CAMs) \cite{selvaraju2017grad} that reveal sensitive regions of an image influencing classification decisions. However, these methods are known to be noisy and often imprecise \cite{adebayo2018sanity}. Recent work includes the ProtoPNet model \cite{protopnet}, which first learns a set of template image patches (or prototypes) for each class during training, and then uses those templates to both predict and explain the class label of a test image. These methods suffer from two drawbacks. First, they do not allow for structured knowledge to guide the learning of interpretable features and hence are not designed to produce results that are \textit{biologically meaningful}. Second, they are mostly developed for classification problems and cannot be directly applied to image generation or translation problems.

 \noindent \textbf{Disentangling ML Features:}
 Another related line of research involves disentangling the feature space of deep learning models to align the disentangled features with target ``concepts.'' This includes the approach of ``Concept whitening'' (CW) \cite{chen2020concept}, where the latent space of a classification model is whitened (i.e., normalized and decorrelated) such that the features along every axis of the latent space corresponds to a separate class. Another approach in this area is that of Latent Space Factorization (LSF) \cite{li2020latent}, where the latent space of an autoencoder is linearly transformed using matrix subspace projections to partition it into features aligned with target concepts (or attributes) and features that do not capture attribute information. Note that our proposed \phylonn{} model can also be viewed as a latent space disentanglement technique, where the disentangled segments of the learned Imageome correspond to different ancestry levels of the phylogeny. We thus use CW and LSF as baselines in our experiments to test if it is possible to discover evolutionary traits just by disentangling the latent space using species classes as orthogonal concepts, without using the structured knowledge of how species are related to one another in the phylogeny. 

 \noindent \textbf{Knowledge-Guided ML:}
 KGML is an emerging area of research that aims to integrate scientific knowledge in the design and learning of ML models to produce generalizable and scientifically valid solutions \cite{karpatne2022knowledge}. Some examples of previous research in KGML include modifying the architecture of deep learning models to capture known forms of symmetries and invariances \cite{wang2020incorporating,anderson2019cormorant}, and adding loss functions that constrain the model outputs to be scientifically consistent even on unlabeled data \cite{raissi2019physics,daw2017physics}. In biology, KGML methods have been developed for species classification that leverage the knowledge of taxonomic grouping of species \cite{hgnn, dos2019improving}. KGML methods have also been developed for generative modeling of images using domain knowledge available as knowledge graphs or ontologies \cite{qi2019ke,garozzo2021knowledge}. In contrast to these prior works, we focus on structuring the embedding space of neural networks using tree-based knowledge (i.e., phylogeny) to enable the discovery and analysis of novel evolutionary traits automatically from images.

\section{Proposed Approach: \phylonn{}} \label{sec:proposed}


\begin{figure}[t]
  \centering
  \includegraphics[width=1.0\linewidth]{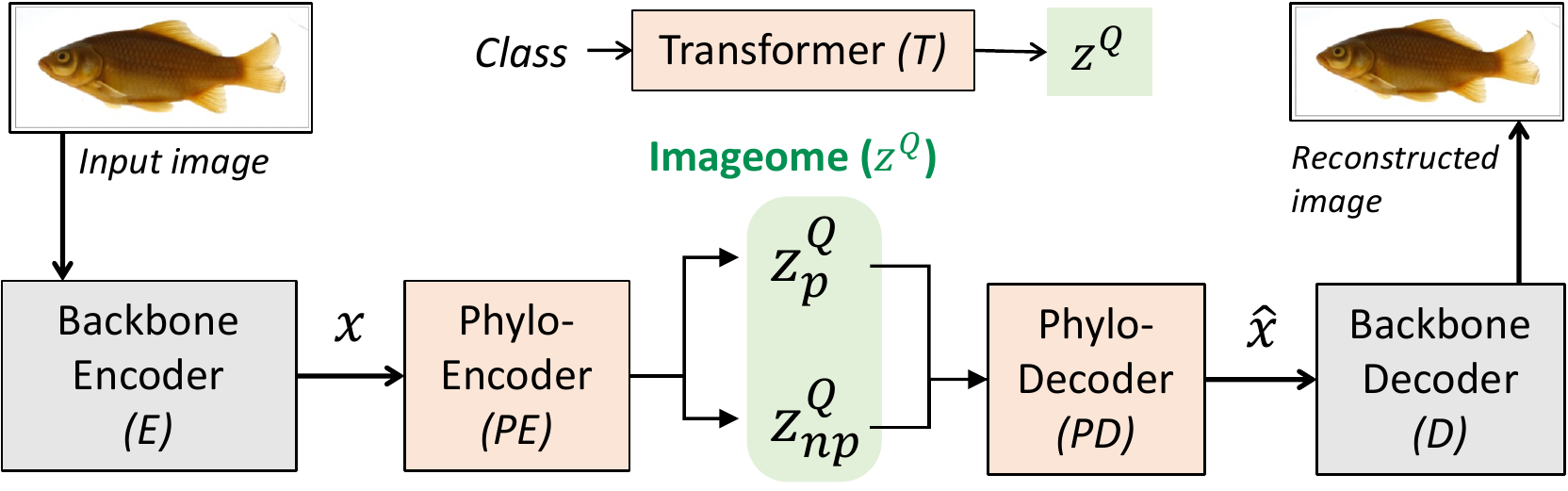}
  \vspace{-4ex}
  \caption{Overview of proposed \phylonn{} model architecture.}
  \label{fig:method}
  \vspace{-4ex}
\end{figure}

We consider the problem of discovering novel (or ``unknown'') evolutionary traits automatically from images without using any trait labels or knowledge of how the unknown traits correspond to known concepts in a knowledge graph or ontology. 
 We only use the ``distant'' supervision of how these unknown traits have evolved over time and are shared across species, available in the form of the phylogenetic tree. 
\cref{fig:method} provides an overview of our proposed \phylonn{} model. Our method can operate on the latent space of any backbone encoder model $\encoder{}$ that takes in images as input and produces continuous feature maps $\imagein{}$ as output. 
 There are three computing blocks in \phylonn{} as shown in \cref{fig:method}. The first block, Phylo-Encoder ($\phyloencoder{}$), takes continuous feature maps  $\imagein{}$ as input and generates quantized feature sequences (or Imageomes) as output. Imageome sequences $\zq{}$ comprise of two \textit{disentangled} segments: $\zphyloq{}$, which captures phylogenetic information (p) at varying ancestry levels, and $\znonattrq{}$, which captures non-phylogenetic  information (np) that is still important for image reconstruction but is unrelated to the phylogeny. The second block, Phylo-Decoder ($\phylodecoder{}$), maps the Imageome sequences back to the space of feature maps $\imageout{}$, such that $\imageout{}$ is a good reconstruction of $\imagein{}$. We then feed $\imageout{}$ into a backbone decoder model $\decoder{}$ that reconstructs the original image. Note that in the training of $\phyloencoder{}$ and $\phylodecoder{}$ models, both the backbone models $\encoder{}$ and $\decoder{}$ are kept frozen, thus requiring low training time. \phylonn{} can thus be plugged into the latent space of any powerful encoder-decoder framework. The third block of \phylonn{} is a transformer model $\transformer{}$ that takes in the species class variable as input, and generates a distribution of plausible Imageome sequences corresponding to the class as output. These sequences can be fed to the $\phylodecoder{}$ model to generate a distribution of synthetic images. 
In the following, we provide details on each of the three blocks of \phylonn{}.


 \begin{figure}[t]
  \centering
  \includegraphics[width=0.55\linewidth]{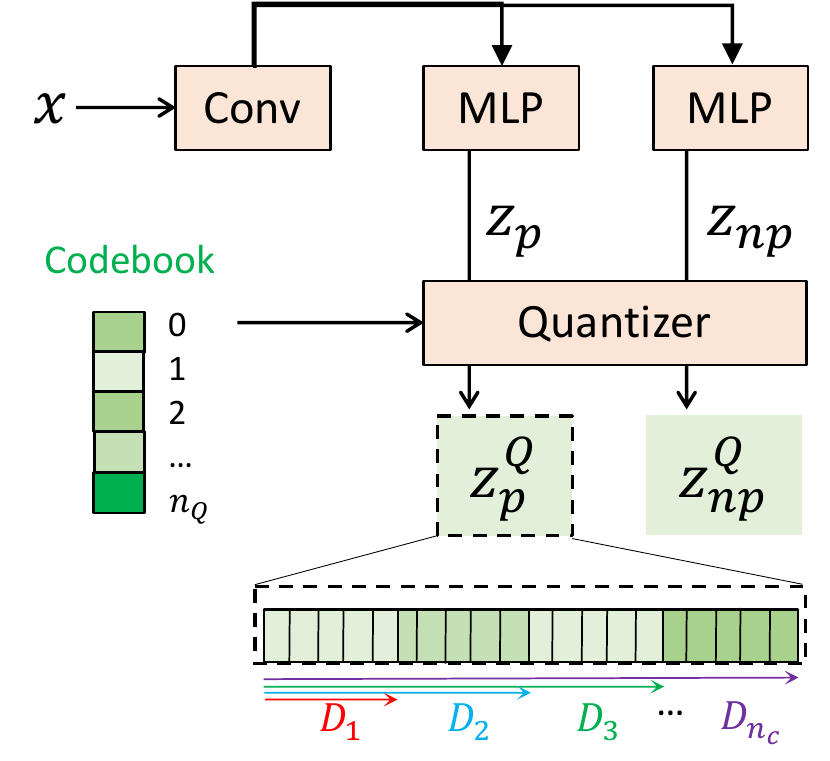}
  \vspace{-2ex}
  \caption{Detailed view of the Phylo-Encoder block.}
  \label{fig:phylo-enc}
  \vspace{-4ex}
\end{figure}

 \subsection{Phylo-Encoder (PE) Block}
\cref{fig:phylo-enc} shows the sequence of operations that we perform inside the $\phyloencoder{}$ block. We first apply a convolutional layer on $\imagein{}$ to produce feature maps of size $(\hin \times \win \times \chin)$, where $\chin$ is the number of channels. We split these $\chin{}$ feature maps into two sets. The first $\chphylo{}$ maps are fed into an MLP layer to learn a global set of feature vectors $\zphylo{}$ capturing phylogenetic information. The size of $\zphylo{}$ is kept equal to $(\nlevelsphylo{}\codesperlevel{}\times \embeddingsize{})$, where $\nlevelsphylo{}$ is the number of phylogeny levels, $\codesperlevel{}$ is the number of feature vectors we intend to learn at every phylogeny level, and $\embeddingsize{}$ is the dimensionality of feature vectors. Similarly, the remaining $\chin - \chphylo$ maps are fed into an MLP layer to produce a set of feature vectors $\znonattr{}$ capturing non-phylogenetic information of size $(\nonattrcodes{} \times \embeddingsize{})$.

\noindent \textbf{Vector Quantization:} Both $\zphylo{}$ and $\znonattr{}$ are converted to \textit{quantized} sequences of feature vectors, $\zphyloq{}$ and $\znonattrq{}$, respectively, using the approach developed in VQVAE \cite{vqvae}. The basic idea of this quantization approach is to learn a set (or codebook) of $\ncodebook{}$ distinct feature vectors (or codes), such that every feature vector in $\zphylo{}$ and $\znonattr{}$ is replaced by its nearest counterpart in the codebook. This is achieved by minimizing the \textit{quantization loss}, $\lquant = |\z{} - \zq{}|$. The advantage of working with quantized vectors is that every feature vector in $\zphyloq{}$ and $\znonattrq{}$ can be referenced just by its location (or index) in the codebook. This allows for faster feature manipulations in the space of discrete code positions than continuous feature vectors.

\label{ss:phyloloss}
\noindent \textbf{Using phylogenetic knowledge in $\zphyloq{}$:} Here, we describe our approach to ensure that the quantized feature sequence $\zphyloq{}$ contains phylogenetic information. Note that $\zphyloq{}$ contains $\nlevelsphylo{}$ sub-sequences of length $\codesperlevel{}$, where every sub-sequence corresponds to a different ancestry level in the phylogeny. While the first sub-sequence $S_1$ should ideally capture information contained in $\imagein{}$ that is necessary for identifying ancestor nodes at level 1 of the phylogeny, $S_2$ should contain additional information that when combined with $S_1$ is sufficient to identify the correct ancestor node of $\imagein{}$ at level 2. In general, we define the concept of a \textit{Phylo-descriptor} $D_i = \{S_1, S_2, \ldots, S_i\}$ of $\imagein{}$ that contains the necessary information for identifying nodes at level $i$ (see \cref{fig:phylo-enc}). We feed $D_i$ to an MLP layer that predicts the class probabilities of nodes at level $i$, which are then matched with the correct node class of $\imagein{}$ at level i, $c_i(\imagein{})$, by minimizing the following \textit{phylogeny-guided loss}, 
$\lphylo{}$: 

\begin{equation}
    \lphylo = \sum_{i=0}^{\nlevelsphylo} \beta_i \CE(\text{MLP}_i(D_{i}(\imagein{})), c_{i}(\imagein{})),
\label{eq:phyloloss}
\end{equation}
where $\CE$ is the cross-entropy loss and $\beta_i$ is the weighting hyper-parameter for level $i$.

\noindent \textbf{Disentangling $\zphyloq{}$ and $\znonattrq{}$:}
While minimizing $\lphylo{}$ guides the learning of $\zphyloq{}$ to contain phylogenetic information, we still need a way to ensure that $\znonattrq{}$ focuses on complementary features and does not contain phylogenetic information. 
To achieve this, we first apply an orthogonal convolution loss $\orthogconv{}$ (originally proposed in \cite{Wang_2020_CVPR}) to the convolutional layer of Phylo-Encoder, to constrain the $\chin{}$ convolutional kernels to be orthogonal to each other. To further ensure that $\znonattrq{}$ has no phylogenetic information, we also employ an {adversarial training} procedure to incrementally remove phylogenetic information from $\znonattrq{}$. In particular, we apply an MLP layer $\advmlp{}$ on $\znonattrq{}$, and then train the parameters of $\advmlp{}$ to minimize the following \textit{adversarial loss}:
\begin{equation}
    \advloss = \sum_{i=0}^{\nlevelsphylo} \beta_i \CE(\text{MLP}_i(\advmlp{}(\znonattrq{}(\mathbf{x}))), c_{i}(\mathbf{x})),
\label{eq:advloss}
\end{equation}
This is aimed at training $\advmlp{}$ to detect any phylogenetic information contained in $\znonattrq{}$. Simultaneously, we train the rest of \phylonn{}'s parameters to maximize $\advloss{}$, such that $\znonattrq{}$ becomes irrelevant for the task of identifying nodes in the phylogeny and only contains non-phylogenetic information.

\section{Phylo-Decoder (PD) Block}
The goal of the PD block is to convert the space of Imageome sequences, $\zq{} = \{\zphyloq{}, \znonattrq{}\}$, back to the space of original feature maps, $\imagein{}$. The sequence of operations in $\phylodecoder{}$ is almost a mirror image of those used in $\phyloencoder{}$. We first pass $\zphyloq{}$ and $\znonattrq{}$ through two MLPs, and then concatenate their outputs to create feature maps of size $(\hin \times \win{} \times \chin{})$. These feature maps are then fed into a convolutional layer to produce $\imageout{}$. Minimizing the reconstruction loss, $\lrec = |\imageout{} - \imagein{}|$, ensures that $\imageout{}$ is a good approximation of $\imagein{}$.
Finally, $\phyloencoder{}$ and $\phylodecoder{}$ are jointly trained using a weighted summation of all the losses mentioned above.

\subsection{Transformer (T) Block} 
\label{ss:transformer}
Once $\phyloencoder{}$ and $\phylodecoder{}$ are trained, we can extract Imageome sequences $\zq{}$ for every image in the training set. The goal of the Transformer block is to learn the patterns of codes in the extracted Imageome sequences of different classes (e.g., species class or ancestor node class), and use these patterns to generate synthetic Imageome sequences for every class. 
To achieve this task, we follow the approach used by VQGAN \cite{esser2021taming} and train a GPT transformer model \cite{radford2019language} $\transformer{}_i$ to generate plausible sequences of $\zq{}$ for every node class at level $i$. The generated Imageome sequences  can then be converted into synthesized specimen images using $\phylodecoder{}$ and $\decoder{}$. 

\section{Evaluation Setup}\label{sec:eval}

\subsection{Data} \label{sec:data}

We used a curated dataset of teleost fish images from five ichthyological research collections that participated in the Great Lakes Invasives Network Project \href{https://greatlakesinvasives.org/portal/index.php}{(GLIN)}. After obtaining the raw images from these collections, we handpicked a subset of about $11,000$ images and pre-processed them by resizing and appropriately padding each image to be of a $256\times256$ pixel resolution. Finally, we partitioned the images into a training set and a validation set using an $80-20$ split. See \cref{app:data} for details on data pre-processing.

Our dataset includes images from 38 species of teleost fishes with an average number of $200$ images per species. We  discretized the phylogenetic tree to have $\nlevelsphylo = 4$ ancestry levels, where the last level is the species class. See \cref{app:phylo} for details on phylogeny selection and discretization.

\subsection{Backbone Encoder and Decoder}
Since \phylonn{} can operate on the feature space $\imagein{}$ of any backbone encoder $\encoder{}$ and produce reconstructed feature maps $\imageout{}$ that can be decoded back to images by a corresponding backbone decoder $\decoder{}$, we tried different encoder-decoder choices including pix2pix \cite{pix2pix2017}, ALAE \cite{pidhorskyi2020adversarial}, and StyleGAN \cite{stylegan2}. However, we found VQGAN \cite{esser2021taming} feature maps to produce images of better visual quality than other encoder-decoder models. Hence, we used the embeddings of a base VQGAN encoder $\encoder{}$ as inputs in \phylonn{} for all our experiments. The reconstructed feature maps of \phylonn{} were then fed into a base VQGAN quantizer serving as the backbone decoder $\decoder{}$. Note that while training \phylonn{}, we kept the parameters of the backbone models fixed, thus saving training time and resources. 

\subsection{Baseline Methods}
Since no direct baselines exist for structuring the embedding space of neural networks using tree-based knowledge or discovering novel evolutionary traits from images, we considered the following baselines that are closest in motivation to \phylonn{}:

\noindent \textbf{Vanilla VQGAN \cite{esser2021taming}:} The first baseline that we consider is a vanilla VQGAN model trained to generate and reconstruct images on the fish dataset. By comparing the learned embeddings and generated images of \phylonn{} with vanilla VQGAN, we aim to demonstrate the importance of using biological knowledge to structure the embedding space of neural networks for trait discovery, rather than solely relying on  information contained in data.

\noindent \textbf{Concept whitening (CW) \cite{chen2020concept}:}
For this second baseline, we replaced the last normalization layer in the encoder block of vanilla VQGAN with the concept whitening (CW) module, where we used species class labels as concept definitions. This is intended to evaluate if CW is capable of disentangling the evolutionary traits of species automatically from images without using the phylogeny. The whitened embeddings $\zconceptwhitening{}$ produced by the CW module are fed into the quantizer module of vanilla VQGAN for converting the embeddings to images. While training the CW module, we optimized the whitening and rotation matrices for all concepts every 30 batches. We used the VQGAN's transformer to generate plausible feature sequences $\zconceptwhitening{}$ conditioned on the species label, which are then decoded into specimen images using the VQGAN's decoder.

\noindent \textbf{Latent Space Factorization (LSF) \cite{li2020latent}:}
The third baseline that we considered is the LSF method, which is another approach for feature disentanglement given concept attribute labels. Specifically, we introduced a variational autoencoder (VAE) model between the encoder and the quantization layer of the base VQGAN model. Similar to CW, we used the species class of each image as the concept attribute for factorizing the latent space in LSF. The LSF module was trained to optimize VAE’s KL-divergence loss and recreation loss along with the attribute and non-attribute losses, as originally defined in the LSF method \cite{li2020latent}. 


\section{Results}\label{sec:results}

In the following, we analyze the results of \phylonn{} from multiple angles to assess the quality of its learned embeddings and generated images in comparison with baseline methods. 







\subsection{Validating Species Distances in the Embedding Space} \label{ss:hamming}
In order to evaluate the ability of \phylonn{} to extract novel (or unknown) evolutionary traits from images without using trait labels, we show that distances between species pairs in the embedding space of \phylonn{} are biologically meaningful and are correlated with ground-truth values better than baseline methods. In the following, we  describe the two types of ground-truths used, the approach used for computing distances in the embedding space of comparative methods, and the comparison of correlations with ground-truth values.


\noindent \textbf{Phylogenetic Ground-truth (GT):} The first ground-truth distance between  pairs of species is the \textit{evolutionary distance} between their corresponding nodes in the phylogenetic tree. In particular, for any two species, we can calculate the total sum of edge lengths in the path between their nodes in the phylogenetic tree. The longer the path, the more distant the species are on the evolutionary scale. Hence, if \phylonn{} indeed captures evolutionary traits in its embedding space, we would expect it to show higher correlations with evolutionary distances computed from the phylogeny as compared to baselines.
We applied min-max scaling of evolutionary distances so that they range from 0 to 1.


\noindent \textbf{Morphological Ground-truth (GT):}
Another type of ground-truth distance between species was computed based on measurements of known morphological traits obtained from the FishShapes v1.0 dataset \cite{price2022}, which contains expert-measured traits known to carry evolutionary signals, defined and collected using traditional methods that are subjective and labor-intensive. 
We specifically used 8 functionally relevant traits from this dataset for every fish species. Some species were not available in this dataset, so when possible, either the closest relative was substituted or the species was dropped.
The species were then matched to a time-calibrated phylogeny of fishes \cite{rabosky2018, chang2019} and the log-transformed measurements were rotated with phylogenetically-aligned components analysis (PACA) \cite{collyer2021}, which rotates the traits to the axis with the highest level of phylogenetic signal. After correcting for overall size and allometry, the principal components of PACA were used to compute the Mahalonobis distance between every species-pair, using a covariance matrix proportional to the evolutionary rate matrix. See \cref{app:morph_dist} for details on PACA calculations. 

\noindent \textbf{Computing Embedding Distances:}
To compute pair-wise distances in the embedding space of \phylonn{}, we first compute the probability distributions (or histograms) of quantized codes at every position of the Imageome sequence (i.e., $\zphyloq{}$ and $\znonattrq{}$) in the test images for every species. We then compute the \textit{Jensen-Shannon (JS) divergence} \cite{menendez1997jensen} between the probability distributions of codes at a pair of species to measure the dissimilarity of their learned  embeddings. We adopt a similar approach for computing the JS-divergence of species-pairs in the quantized feature space of vanilla VQGAN. For baseline methods that operate in continuous feature spaces (CW and LSF), we first calculate the mean feature vector for every species and then compute the \textit{cosine distance} ($1 ~- $ \textit{cosine similarity}) of vectors for a pair of species. For both metrics, JS-divergence and cosine distance, a value closer to 0 represents higher similarity.


\begin{figure}[t]
   \centering
  \begin{subfigure}{0.47\columnwidth}
  \includegraphics[width=\textwidth]{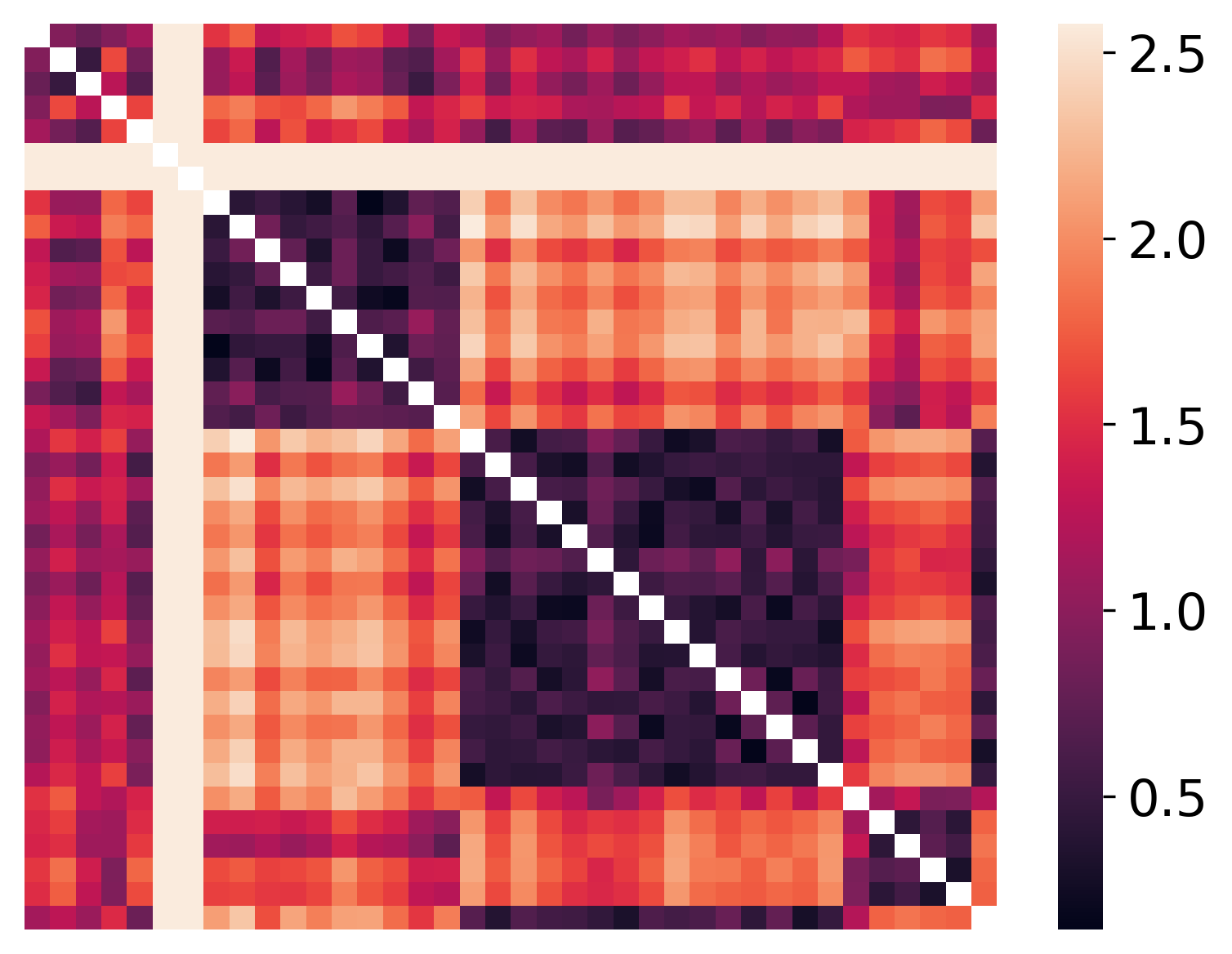}
  \caption{Morphological ground truth}
  \end{subfigure}
  \hfill
  \begin{subfigure}{0.47\columnwidth}
  \includegraphics[width=\textwidth]{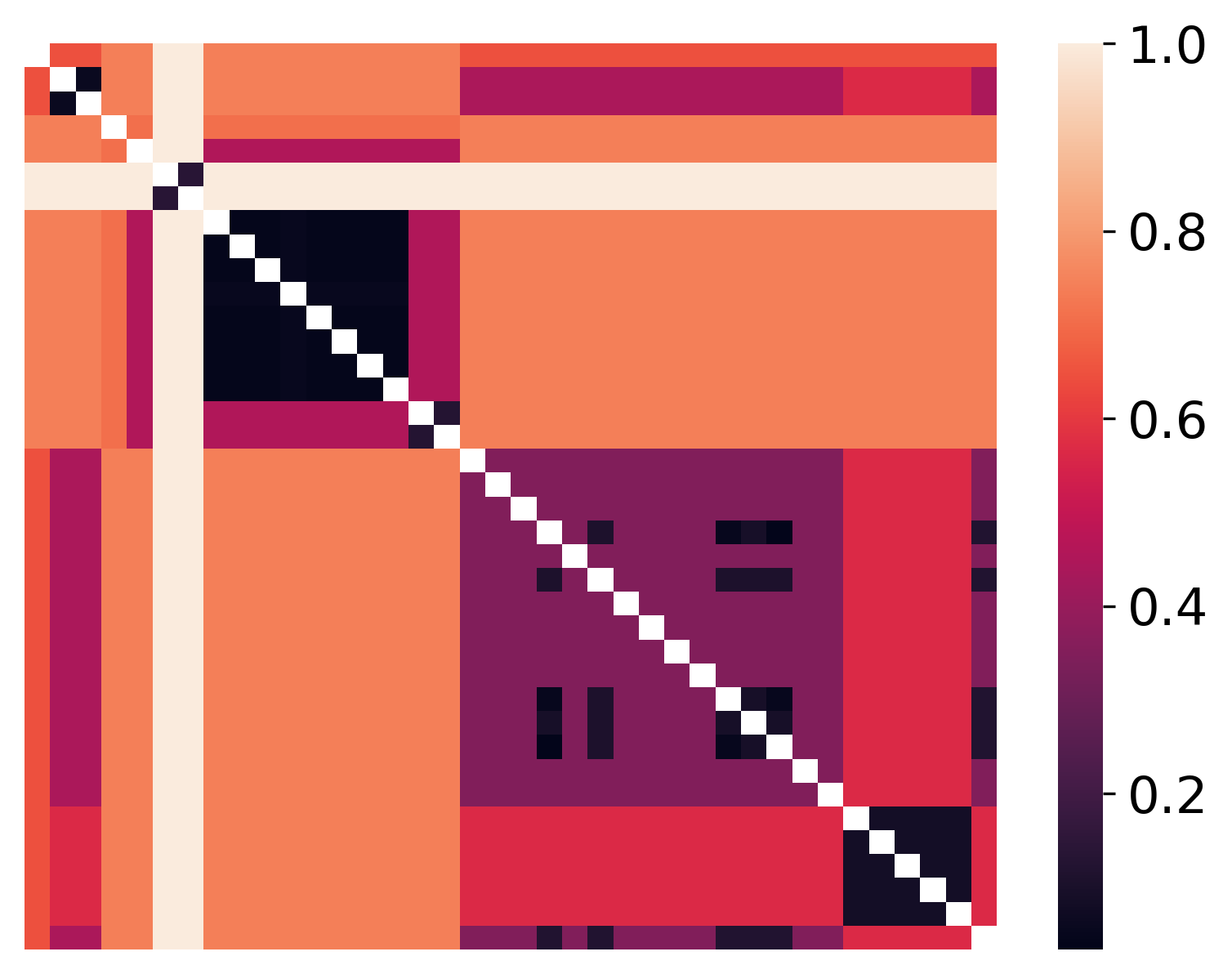}
  \caption{Phylogenetic ground truth} 
  \end{subfigure} 
  \begin{subfigure}{0.47\columnwidth} 
  \includegraphics[width=\textwidth]{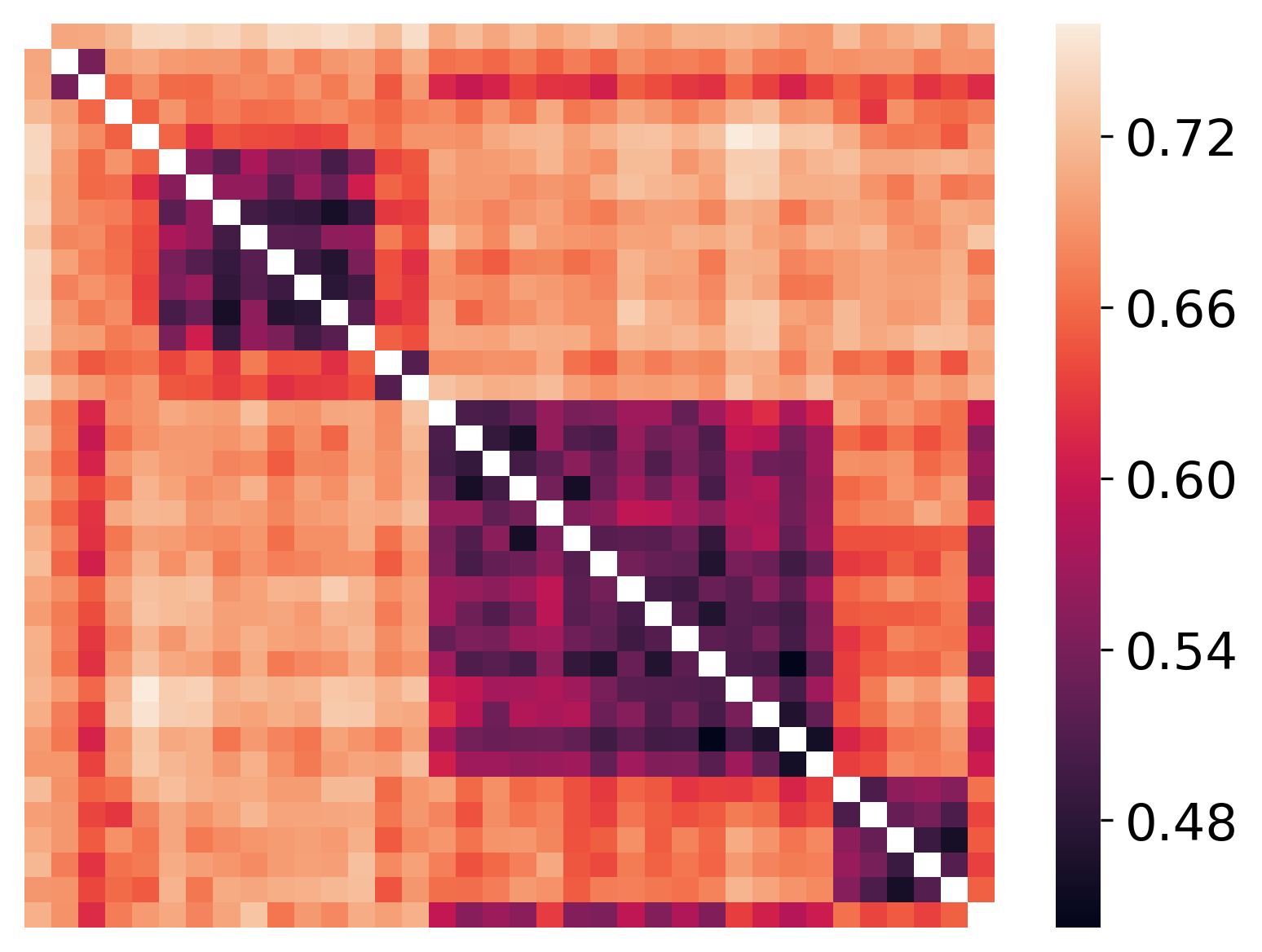} 
  \caption{JS-divergence for \phylonn{}'s phylogenetic embedding} 
  \end{subfigure}  
  \hfill 
  \begin{subfigure}{0.47\columnwidth} 
  \includegraphics[width=\textwidth]{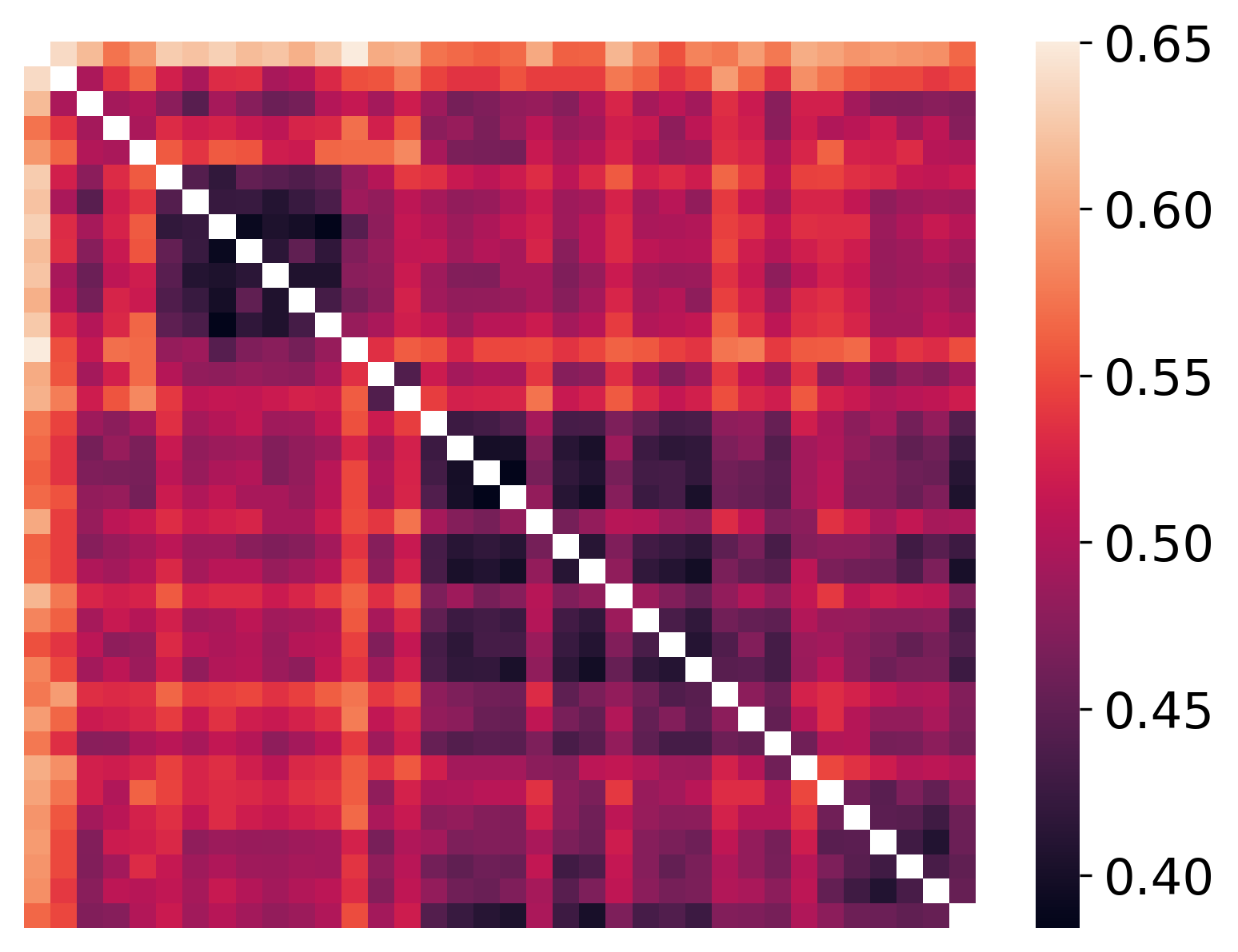} 
  \caption{JS-divergence for \phylonn{}'s non-phylogenetic embedding} 
  \end{subfigure}
  \begin{subfigure}{0.47\columnwidth} 
  \includegraphics[width=\textwidth]{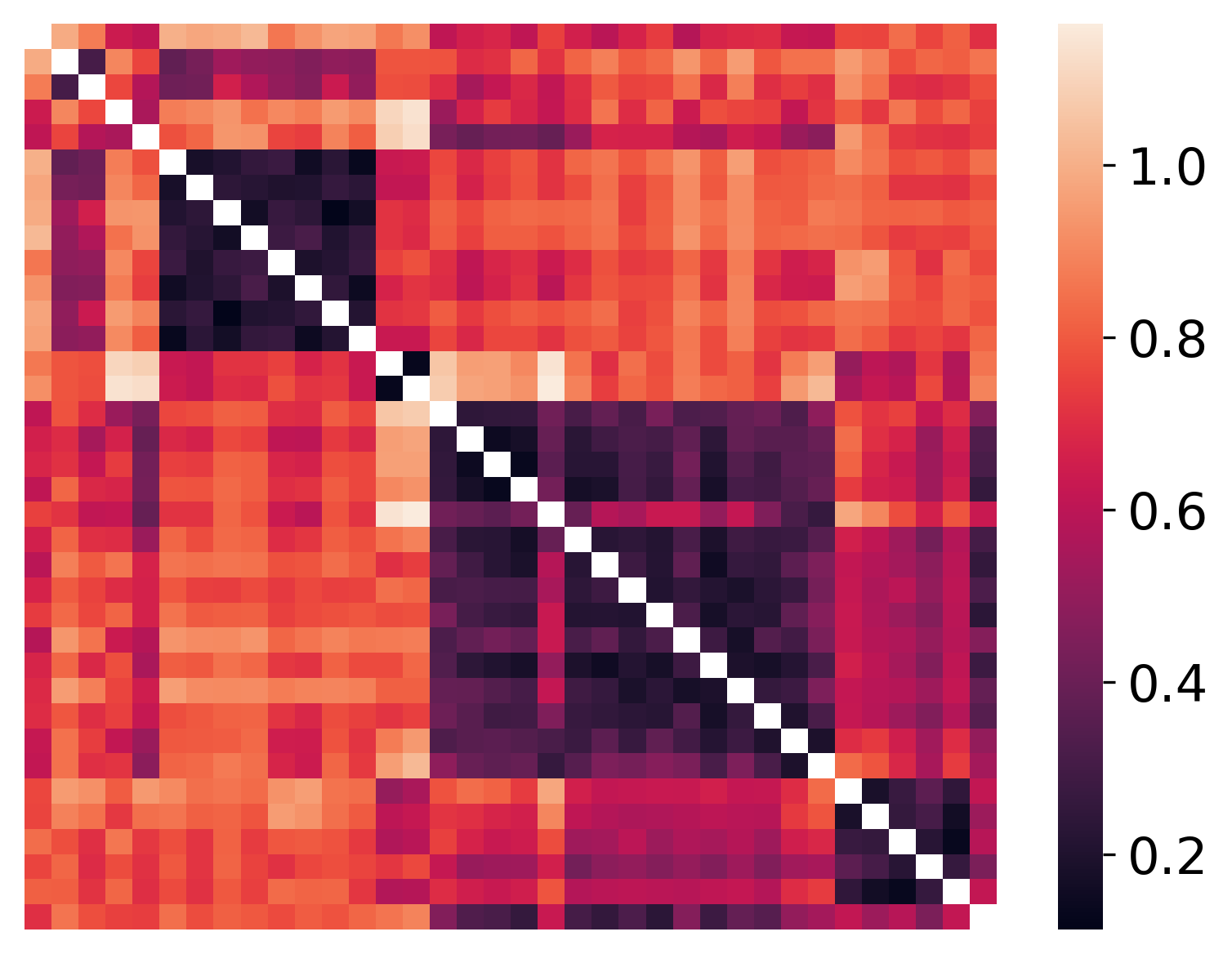} 
  \caption{Cosine distance for CW's \cite{chen2020concept} embedding} 
  \end{subfigure}  
  \hfill 
  \begin{subfigure}{0.47\columnwidth} 
  \includegraphics[width=\textwidth]{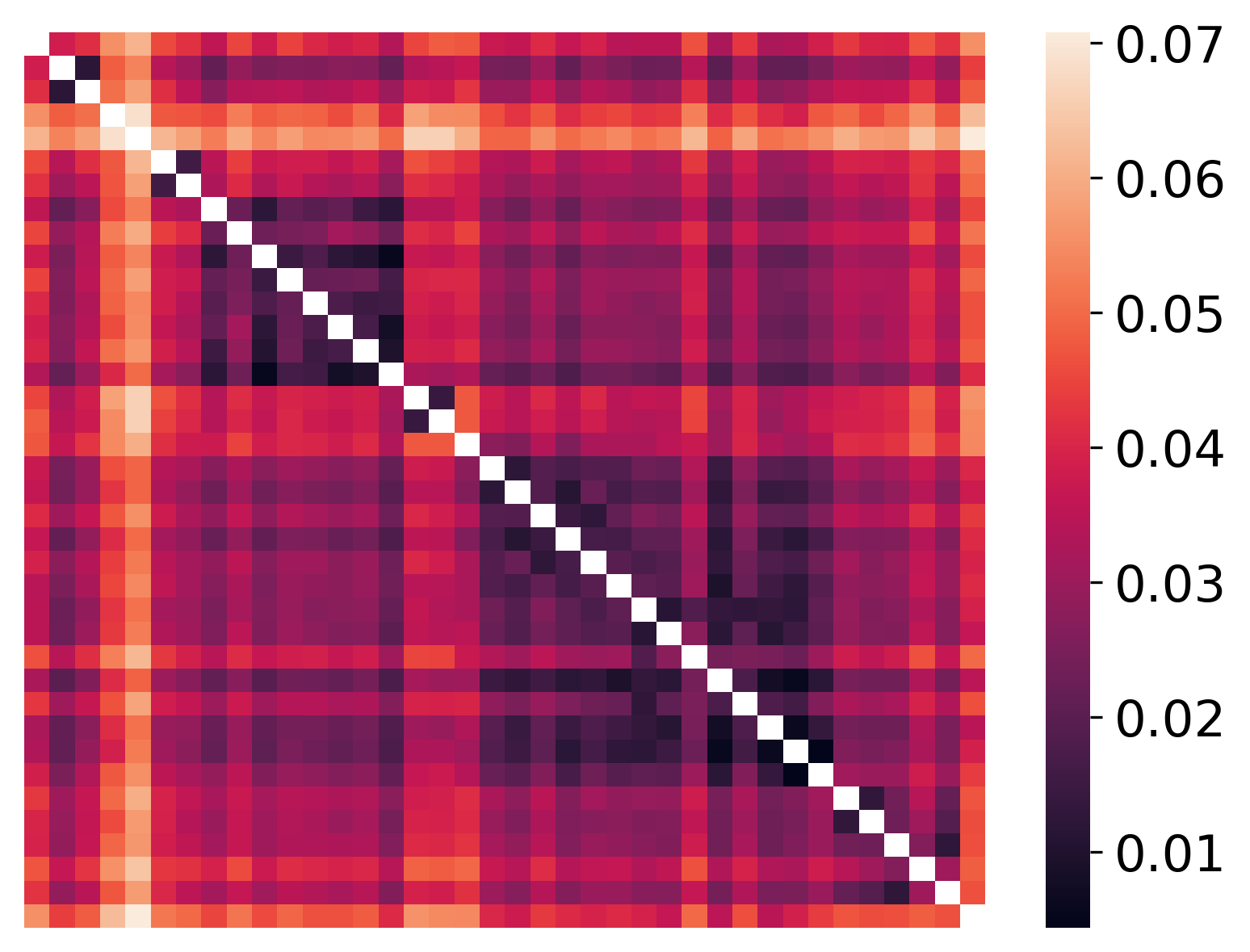} 
  \caption{Cosine distance for LSF's \cite{li2020latent} embedding} 
  \end{subfigure}
  \begin{subfigure}{0.47\columnwidth} 
  \includegraphics[width=\textwidth]{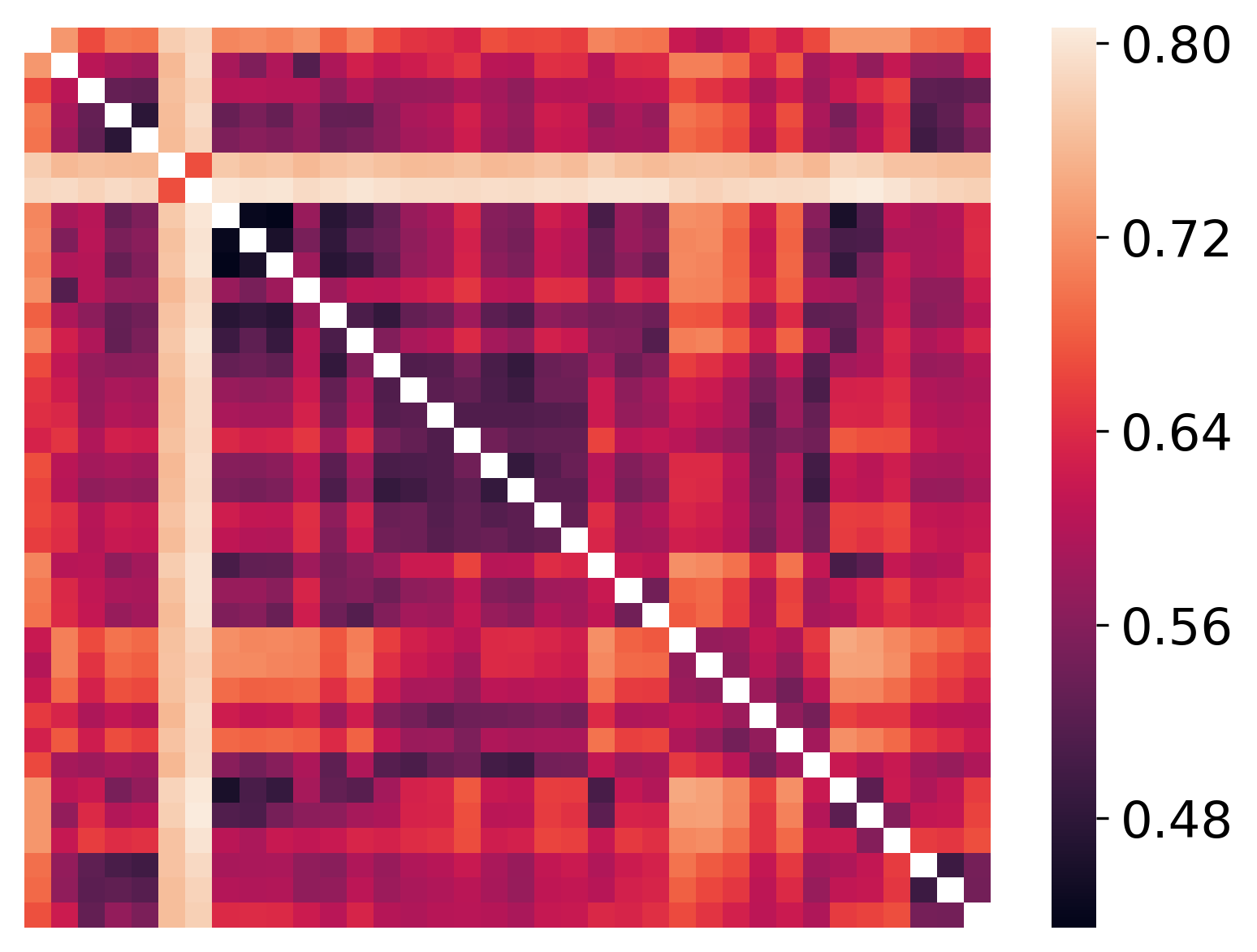} 
  \caption{JS-divergence for Vanilla VQGAN's \cite{esser2021taming} embedding} 
  \end{subfigure}  
  \hfill 
  \caption{Comparing embedding distance matrices of methods with morphological and phylogenetic ground-truths.}
  \label{fig:distances}
  \vspace{-4ex}
\end{figure}

\noindent \textbf{Comparing Correlations with Ground Truth:}
\cref{fig:distances}(a) and \cref{fig:distances}(b) show the pair-wise species distance matrices for morphological and phylogenetic GTs, respectively. Note that the rows and columns of  all matrices in \cref{fig:distances} are species  ordered according to their position in the phylogeny (see \cref{app:phylo} for details) and the diagonal values (correlation with self) are removed so that they do not affect the colormap scale. We can see that both ground-truths show a similar clustering structure of species.
However, there are differences too; while phylogenetic GT is solely based on phylogeny, the morphological GT uses both the phylogeny and information about ``known'' traits.  
\cref{fig:distances}(c) and \cref{fig:distances}(d) show the JS-divergences among species computed separately for the two disentangled parts of PhyloNN's embeddings ($\zphyloq{}$ and $\znonattrq{}$). We can see that the embeddings containing phylogenetic information show a similar clustering structure of distances as the GT matrices, in contrast to the non-phylogenetic embeddings. This shows the ability of \phylonn{} to disentangle features related to phylogeny from other unrelated features. \cref{fig:distances} also shows the embedding distance matrices of the baseline methods, which are not as visually clean as \phylonn{} in terms of matching with the GT matrices.


To quantitatively evaluate the ability of \phylonn{} to match with GT distances, we compute the Spearman correlation between the GT distance matrices and embedding distance matrices for different methods as shown in \cref{tab:correlation}. We can see that \phylonn{} shows higher correlations at the species level with both  GTs. Furthermore, since \phylonn{} learns a different descriptor for every ancestry level in contrast to baseline methods that learn a flat representation, we can also compute \phylonn{}'s distance matrix at any ancestry level and compare it with GT matrices at the same level. \cref{tab:correlation} shows that \phylonn{} shows significantly higher correlations with GT matrices at higher ancestry levels than the species level.


\begin{table}
\centering
\caption{Correlations between GT and embedding distances}
 \label{tab:correlation}
\begin{tabular}{ll|ll}
\toprule
                         &         & Morphological & Phylogenetic \\
\midrule
\multirow{4}{*}{PhyloNN} & level0  &     0.86          &     0.83         \\
                         & level1  &     0.87           &  0.85            \\
                         & level2  &     0.78          &   0.83           \\
                         & species &     0.70          &   0.78           \\
\midrule
\multicolumn{2}{l}{LSF}            &      0.38  &     0.55        \\
\multicolumn{2}{l}{CW}          &      0.70        &     0.67        \\
\multicolumn{2}{l}{vanilla VQGAN}          &    0.31          &    0.24         \\
\bottomrule
\end{tabular}
\vspace{-2ex}
\end{table}



\subsection{Evaluating Species-to-species Image Translations} \label{ss:coderep}

To further assess how well \phylonn{}'s embeddings capture evolutionary traits, we investigate how altering the learned Imageome sequence of an image specimen incrementally in a phylogenetically meaningful ordering affects the observed traits when the altered embeddings are decoded back as an image. To do that, we set up the following experiment. We pick two specimen images from a pair of species. By encoding the two images using \phylonn{}, we obtain their corresponding Imageome encodings, $\zq_1$ and $\zq_2$. We then start to replace the codes in the Imageome sequence $\zq_1$ with the corresponding codes in $\zq_2$ iteratively, until $\zq_1$ transforms completely into $\zq_2$. The order of this iterative replacement is by first replacing the codes representing the non-phylogenetic part of the embedding $\znonattrq_1$, then the part capturing evolutionary information at the earliest ancestry level (level 0), to the next ancestry level (level 1), till we eventually reach the last level of the phylogeny, which is the species level. At the final point, the entire Imageome sequence $\zq_1$ has been replaced with $\zq_2$. This phylogeny-driven ordering of code replacements helps us capture key ``snapshots'' of the species-to-species translation process that are biologically meaningful. In particular, by observing the traits that appear or disappear at every ancestry level of code replacement, we can infer and generate novel hypotheses about the biological timing of trait changes as they may have happened in evolutionary history.

\begin{figure*}[h]
   
   \centering
  \begin{subfigure}[b]{2.0\columnwidth}
  \centering
  \includegraphics[width=\textwidth]{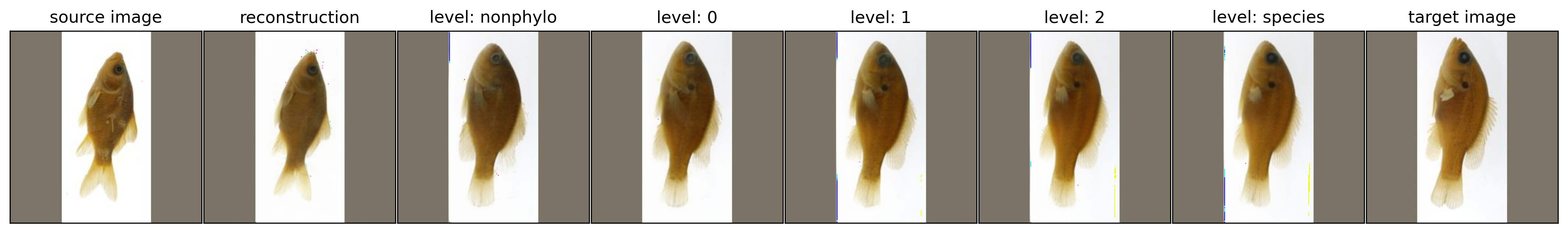}
  \caption{\phylonn{}}
  \end{subfigure}
  \hfill
    \begin{subfigure}[b]{2.0\columnwidth}
  \centering
  \includegraphics[width=\textwidth]{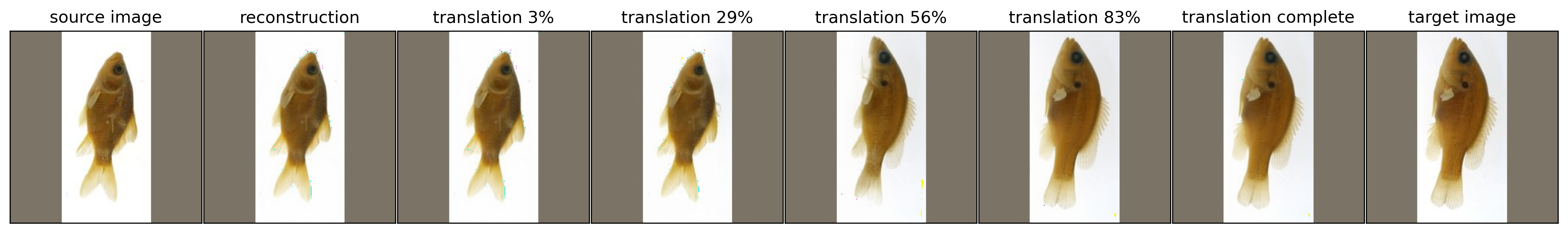}
  \caption{Vanilla VQGAN / CW}
  \end{subfigure}
  \hfill
    \begin{subfigure}[b]{2.0\columnwidth}
  \centering
  \includegraphics[width=\textwidth]{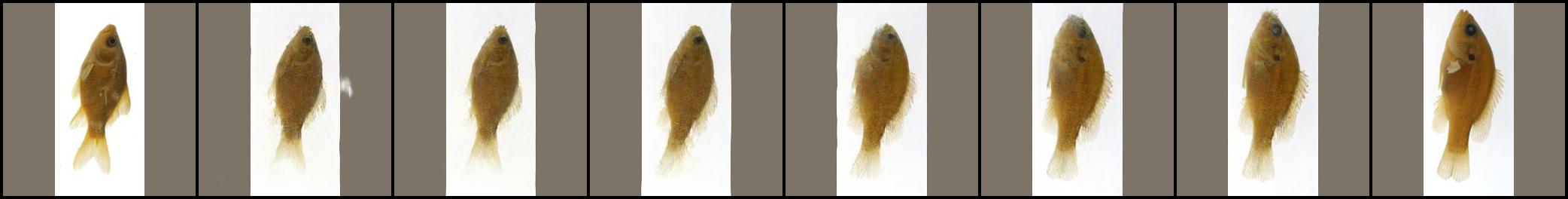}
  \caption{LSF}
  \end{subfigure}
  \hfill
  \caption{Comparing species-to-species image translations from a \textit{Carassius auratus} specimen to a \textit{Lepomis cyanellus} specimen.}
    \label{fig:transitions}

\end{figure*}

\cref{fig:transitions} shows an example of such a translation process between a specimen of the species  \textit{\textit{Carassius auratus}} to a specimen of the species \textit{\textit{Lepomis cyanellus}}. We can see that although the two specimens look similar on the surface, there are several subtle traits that are different in the two species that are biologically interesting. For example, the source species has a V-shaped tail fin (termed caudal fin), while the target species has a rounded caudal fin. By looking at their place of occurrence in the translation process of \phylonn{}, we can generate novel biological hypotheses of whether they are driven by phylogeny or not, and whether they appeared earlier or later in the target species in the course of evolution. For example, we can see that the rounded tail feature of the target species appears right after replacing the non-phylogenetic part of the embedding (see blue circle), indicating that this feature may not be capturing evolutionary signals and instead maybe affected by unrelated factors (e.g., environment). On the other hand, if we observe another fin (termed pectoral fin) that appears on the side of the body just behind the gill cover, compared to the fin's lower position closer to the underside of the specimen in the source image, we can see that it seems to get sharper and compact only in the later levels (it is faintly visible in level 1 but shows up prominently as a white region in level 2, see green circle). This suggests that the change in the position and shape of the pectoral fin occurred later in fish evolution, which is supported by the phylogeny and is in fact the case. Our work opens novel opportunities for generating such biological hypotheses, which can be further investigated by biologists to potentially accelerate scientific discoveries. \cref{fig:transitions} also shows the translations obtained by baseline methods for the same pair of species specimens. We can see that the baselines are mostly performing a smooth interpolation between the source and target images. This is in contrast to the discrete and non-smooth nature of changes observed in the translation of \phylonn{}, which is indeed the desired behavior since the appearance or disappearance of traits at every ancestry level are expected to be orthogonal to those at other levels.
Furthermore, the transition points in the translation process of baseline methods do not correspond to biologically meaningful events as opposed to \phylonn{}.  


\subsection{Generalization to Unseen Species}
\label{ss:unseen}

As \phylonn{} aims to encode specimen images into their corresponding phylogenetic and non-phylogenetic sequences of codes, 
we expect specimens that belong to the same species to largely share the same phylogenetic code in terms of the species descriptor $D_{\nlevelsphylo{}}$, while varying in terms of the non-phylogenetic codes. More generally, specimens belonging to species sharing a common ancestor at phylogenetic level $i$ should largely share the codes with the descriptor at level $i$, $D_{i}$, while varying at the rest of codes. This should also apply for specimens of \textit{unseen} (or newly discovered) species that we have not yet observed in the training set. We posit that by looking at the similarity of codes generated for an unseen species, we should be able to infer its ancestral lineage in terms of the species sampled during training. In other words, by analyzing the distribution of codes generated from the image of an unseen species, we should be able to locate it on the phylogenetic tree and position it to next to the subset of known species (seen during training) that share a common ancestor.  

To quantify this phenomenon, for a given species or ancestor node, we construct two sets of histograms, $\Hphylo{}$ and $\Hnonattr{}$, of sizes $[ \nlevelsphylo{} \times{} \codesperlevel{}]$ and $[ \nonattrcodes{}]$, respectively. Each value in the histograms, $\Hphylo{}^{i,j}$ and $\Hnonattr{}^{k}$, describes the distribution of codes at a certain location in the Imageomes  across all specimens belonging to the species or ancestor node. See \cref{app:hist} for an example.
We then compute the entropy of each sequence location in $\Hphylo{}$ and $\Hnonattr{}$ to measure the ``purity'' of codes used at every location. If the entropy is low for a certain  location, it means only a few possible codes occur at that location, suggesting that those specific codes are key at characterizing the species or ancestor node in question. On the other hand, higher entropy means a variety of codes occur at that location, implying that such a location is not discriminative to the species or ancestor node.
Finally, to compare the code distributions for two species, we use the JS-divergence metric for calculating the difference between two histograms of a sequence location. Similar to \cref{ss:hamming}, such a metric can be aggregated to quantify the coding differences between species-pairs.

To assess  \phylonn{}'s ability to generalize to unseen species, we train it on a subset of the species and then evaluate the quality of the embedding space when the model is introduced to species it has never seen before during training. In our experiment, we chose to train on the same dataset as before while only excluding three species. Once the model is trained, we look at the average JS-divergence distance between these missing species and three other species in the tree. These three other species were selected such that each missing species has one seen species that is close to it phylogenetically (i.e., both species share the same ancestor at the immediate ancestry level) while  others are relatively far from it. 

\cref{tab:jsnunseenphylo} shows the average distance of the phylogenetic codes among the six aforementioned species. We can see that the distance is smallest for each unseen species and its counterpart that shares the same immediate ancestor (shown as the diagonal in the table). This confirms that even though the model has not seen the former species, it was able to characterize it using an Imageome sequence that is significantly closer to that of its seen counterpart than the other species' Imageomes.

While \cref{tab:jsnunseenphylo} highlights the phylogenetic matching in the embedding space at the species descriptor level, $D_{\nlevelsphylo{}}$, \cref{tab:jsnunseenlvl0} does the same but for the descriptor at a distant ancestry level (level 0), i.e., $D_{0}$. Based on the phylogenetic tree we have used in this example, both the \textit{Notropis} and \textit{Noturus} species share the same distant ancestor at that descriptor level. On the other hand, \textit{Lepomis} species does not share that ancestor. Hence, we find that the JS-divergences increase for the \textit{Lepomis} unseen species with seen species that are not \textit{Lepomis} as compared to \cref{tab:jsnunseenphylo}. On the other hand, the JS-divergences decrease for the other two unseen species w.r.t. seen species that are on the off-diagonals of the table. This confirms that $D_{0}$ specifically captures the phylogenetic information of that distant ancestor that is common across \textit{Notropis} and \textit{Noturus} seen and unseen species.
Finally, to confirm that this phylogenetic correlation is mainly constrained only to the phylo-descriptors, we calculate the same distances but using the non-phylogenetic part of the sequences. The result is shown in \cref{tab:jsnunseennonphylo}. We can see that the distances are much closer to each other, implying that the non-phylogenetic embedding is not specialized at differentiating among  different species, and hence cannot be used to phylogenetically categorize the unseen species.

\begin{table}[]
\caption{JS-diveregence of the phylogenetic codes at the species level between unseen and seen species}
 \label{tab:jsnunseenphylo}
\begin{tabular}{p{0.05\linewidth}|p{0.22\linewidth}|p{0.16\linewidth} p{0.16\linewidth} p{0.16\linewidth}}
\toprule
            &             &         \multicolumn{3}{c}{Seen species} \\
\midrule
            &             &     \textit{Notropis nubilus}    &  \textit{Lepomis macrochirus} & \textit{Noturus flavus} \\
\midrule
\multirow{4}{*}{\rotatebox[origin=c]{90}{Unseen species}}  & \textit{Notropis percobromus}  & 0.47  &     0.71          &     0.62         \\
                                 & \textit{Lepomis megalotis}  & 0.73  &    0.43           &  0.72            \\
                                 &  \textit{Noturus miurus} & 0.62  &     0.71          &   0.48           \\
\bottomrule
\end{tabular}
\end{table}

\begin{table}[]

\caption{JS-diveregence of the phylogenetic codes at the earliest ancestral level between unseen and seen species}
 \label{tab:jsnunseenlvl0}
\begin{tabular}{p{0.05\linewidth}|p{0.22\linewidth}|p{0.16\linewidth} p{0.16\linewidth} p{0.16\linewidth}}
\toprule
            &             &         \multicolumn{3}{c}{Seen species} \\
\midrule
            &             &     \textit{Notropis nubilus}    &  \textit{Lepomis macrochirus} & \textit{Noturus flavus} \\
\midrule
\multirow{4}{*}{\rotatebox[origin=c]{90}{Unseen species}}  & \textit{Notropis percobromus}  & 0.26  &     0.81          &     0.50         \\
                                 & \textit{Lepomis megalotis}  & 0.81  &    0.27           &  0.81            \\
                                 &  \textit{Noturus miurus} & 0.52  &     0.80          &   0.31           \\
\bottomrule
\end{tabular}
\end{table}

\begin{table}[]
\caption{JS-diveregence of the non-phylogenetic codes between unseen and seen species}
 \label{tab:jsnunseennonphylo}
\begin{tabular}{p{0.05\linewidth}|p{0.22\linewidth}|p{0.16\linewidth} p{0.16\linewidth} p{0.16\linewidth}}
\toprule
            &             &         \multicolumn{3}{c}{Seen species} \\
\midrule
            &             &     \textit{Notropis nubilus}    &  \textit{Lepomis macrochirus} & \textit{Noturus flavus} \\
\midrule
\multirow{4}{*}{\rotatebox[origin=c]{90}{Unseen species}}  & \textit{Notropis percobromus}  & 0.39  &     0.45          &     0.39         \\
                                 & \textit{Lepomis megalotis}  & 0.46  &    0.36           &  0.48            \\
                                 &  \textit{Noturus miurus} & 0.40  &     0.42          &   0.36           \\
\bottomrule
\end{tabular}
\vspace{-4ex}
\end{table}





\subsection{Assessing the Clustering Quality of the Embedding Space Using t-SNE Plots}\label{ss:tsne} In this section, we  qualitatively assess the quality of generated images by visualizing their embedding space. Visualization tools such as loss landscape visualizations \cite{NEURIPS2018_a41b3bb3} and t-SNE plots \cite{JMLR:v9:vandermaaten08a}, have been frequently used as investigative tools in deep learning in recent years as they help gauge a model's generalization power. To that end, we are interested in understanding how \phylonn{} clusters the embedding space compared to other baselines by analyzing these models' t-SNE plots. 
To construct the t-SNE plot for each model, we iterate through its generated images, encode them, obtain the quantized embedding vector for each image ($\zphyloq{}$ and $\zq{}$ for \phylonn{} and vanilla VQGAN, respectively), and finally create the t-SNE plots. For CW, we use the whitened embeddings $\zconceptwhitening{}$ instead.

\begin{figure}[h]
    
   \centering
  \begin{subfigure}[b]{1.0\columnwidth}
  \centering
  \includegraphics[width=0.42\textwidth]{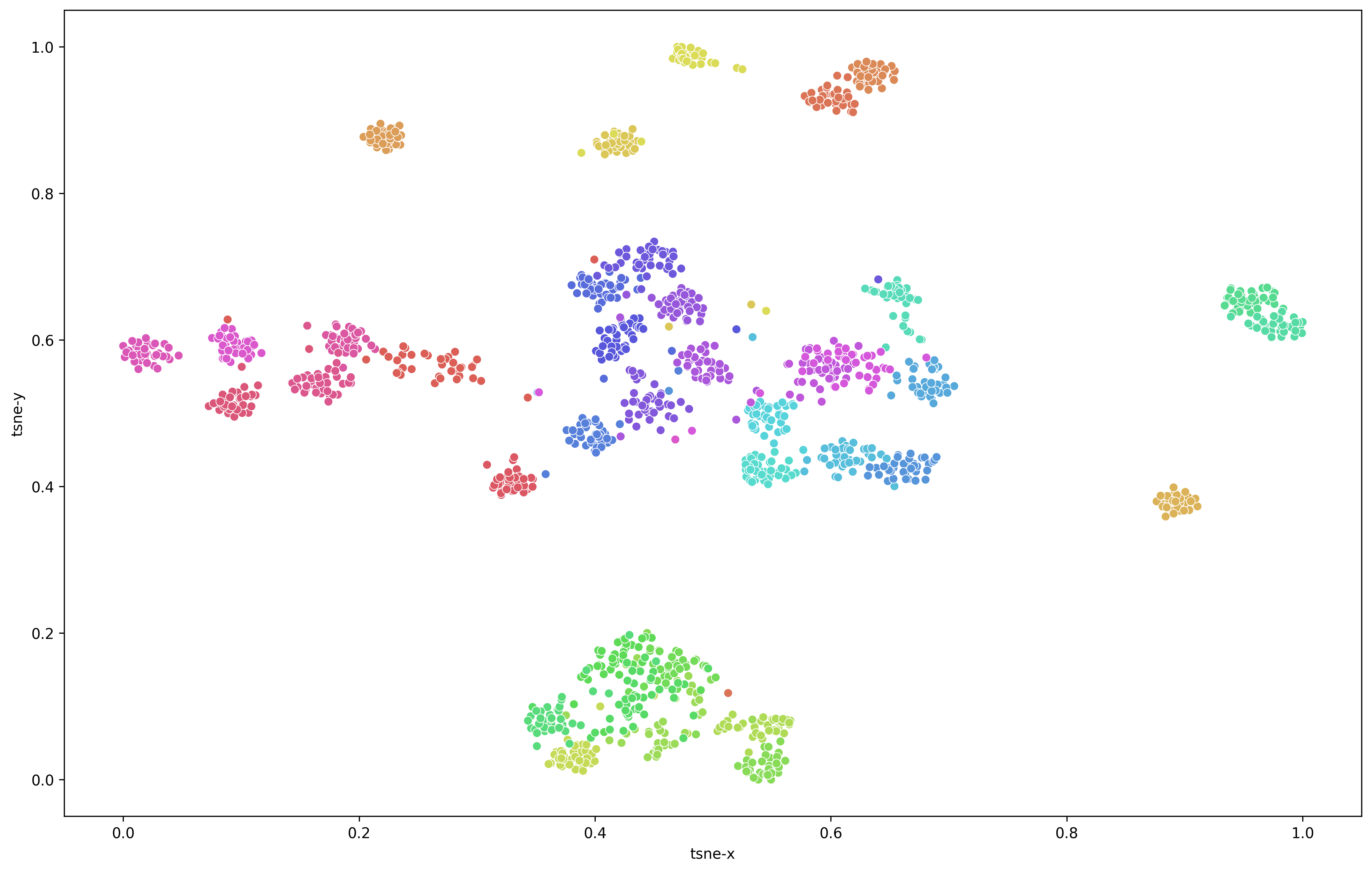}
  \hfill
  \includegraphics[width=0.42\textwidth]{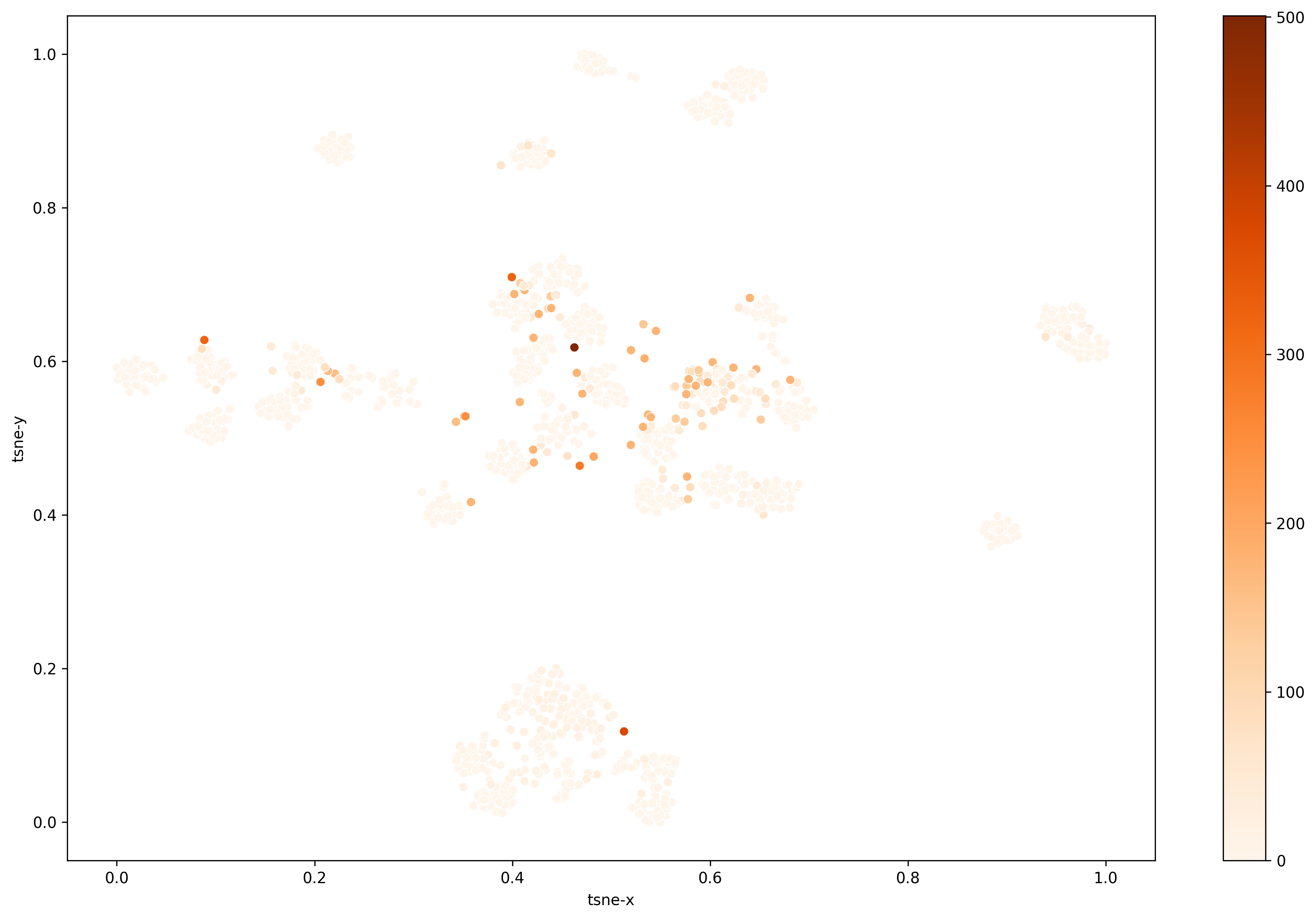}
  \caption{\phylonn{}}
  \vspace{-2ex}
  \end{subfigure} 
  \vskip\baselineskip
  \begin{subfigure}{1.0\columnwidth} 
  \includegraphics[width=0.42\textwidth]{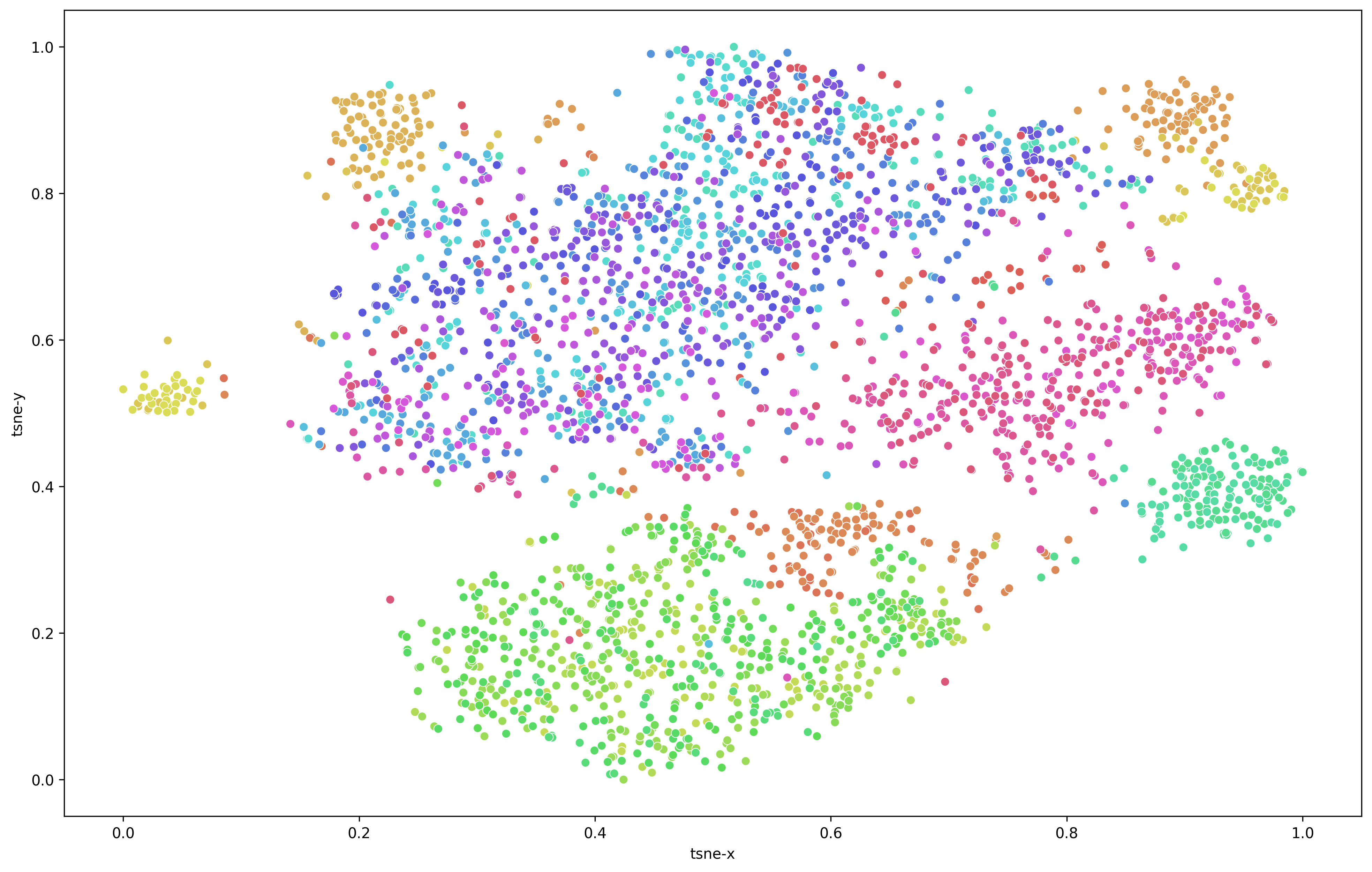} 
  \hfill
  \includegraphics[width=0.42\textwidth]{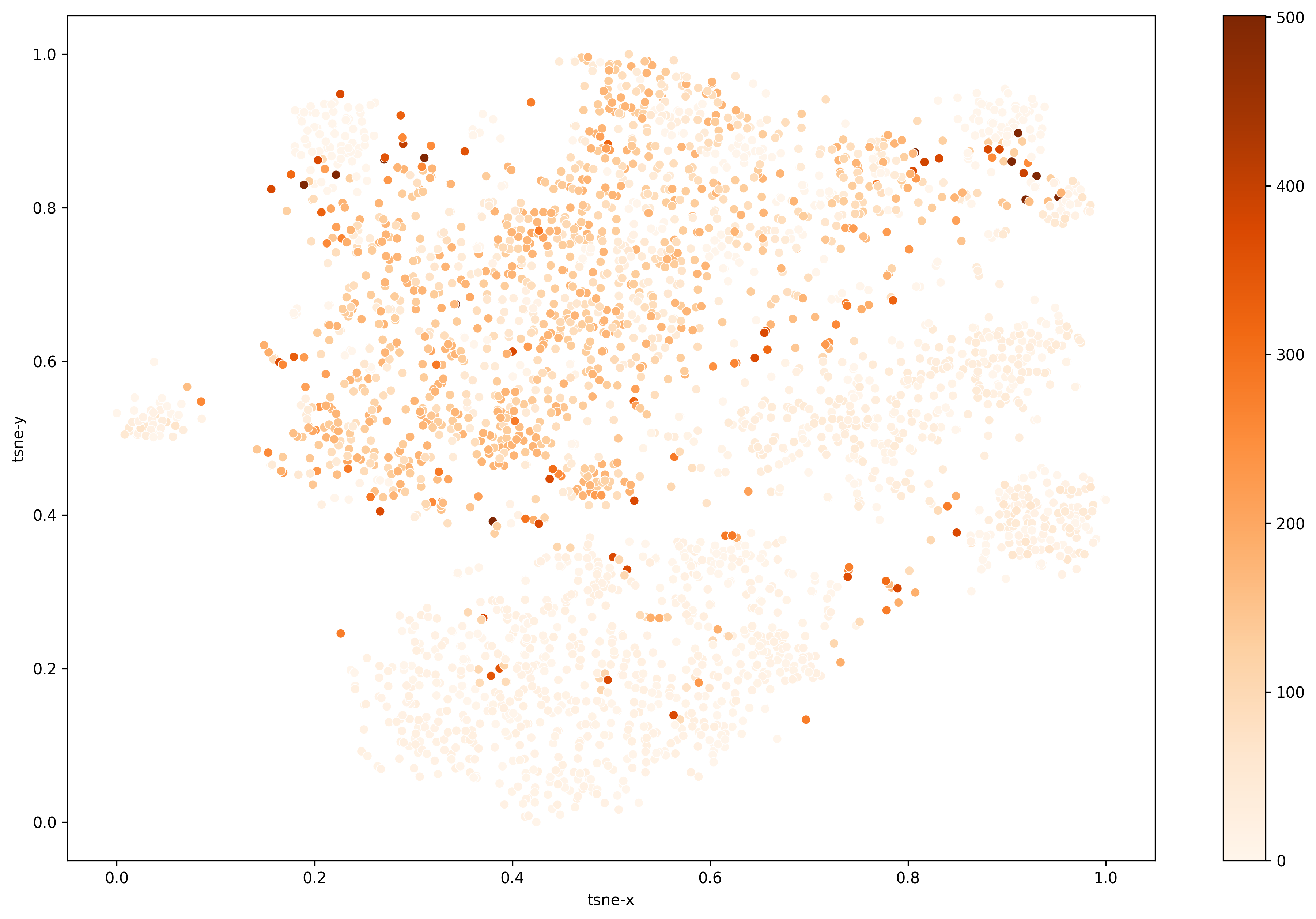} 
  \caption{Vanilla VQGAN} 
  \vspace{-2ex}
  \end{subfigure}
\vskip\baselineskip
  \begin{subfigure}{1.0\columnwidth} 
  \includegraphics[width=0.42\textwidth]{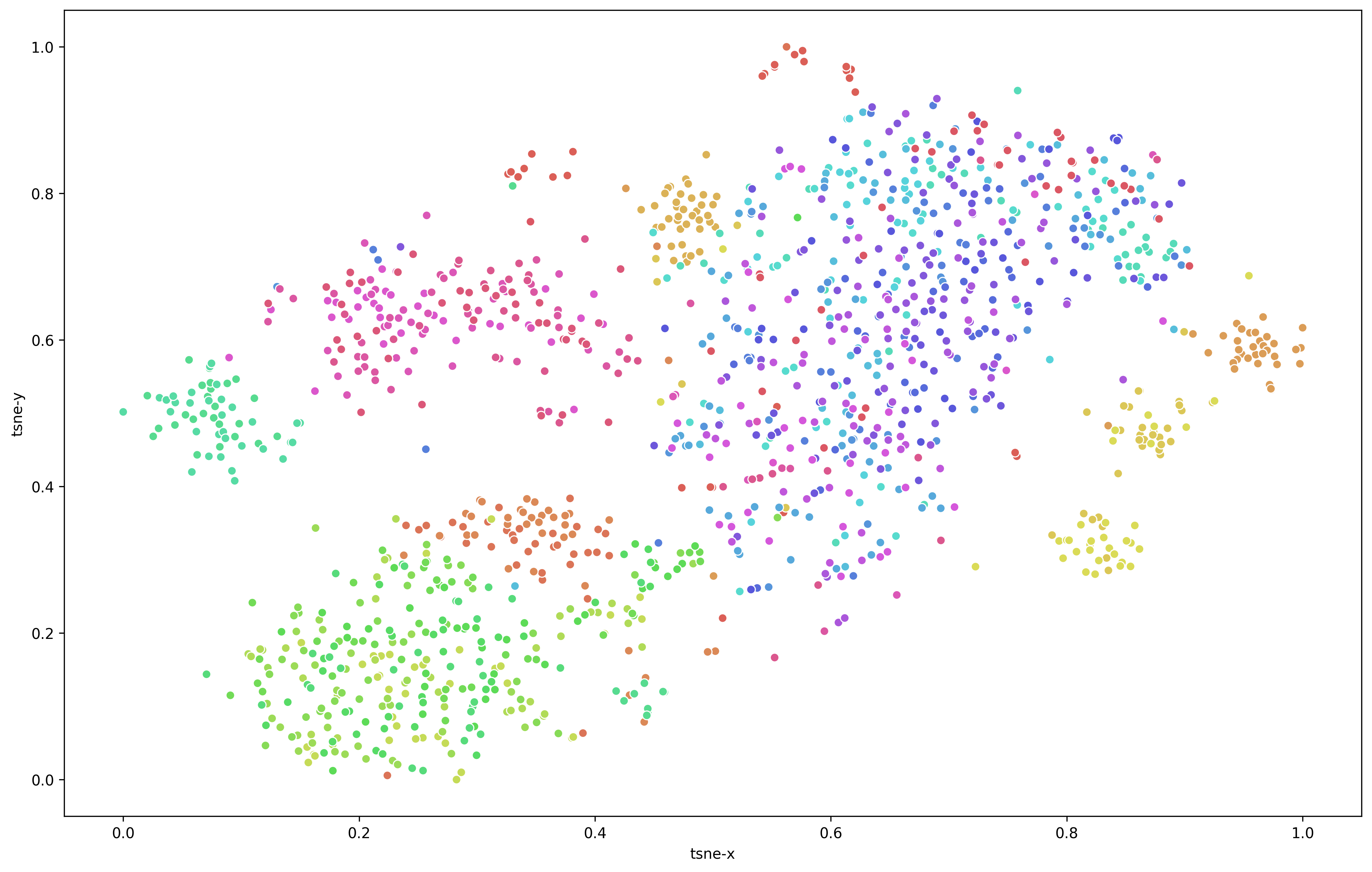} 
  \hfill
  \includegraphics[width=0.42\textwidth]{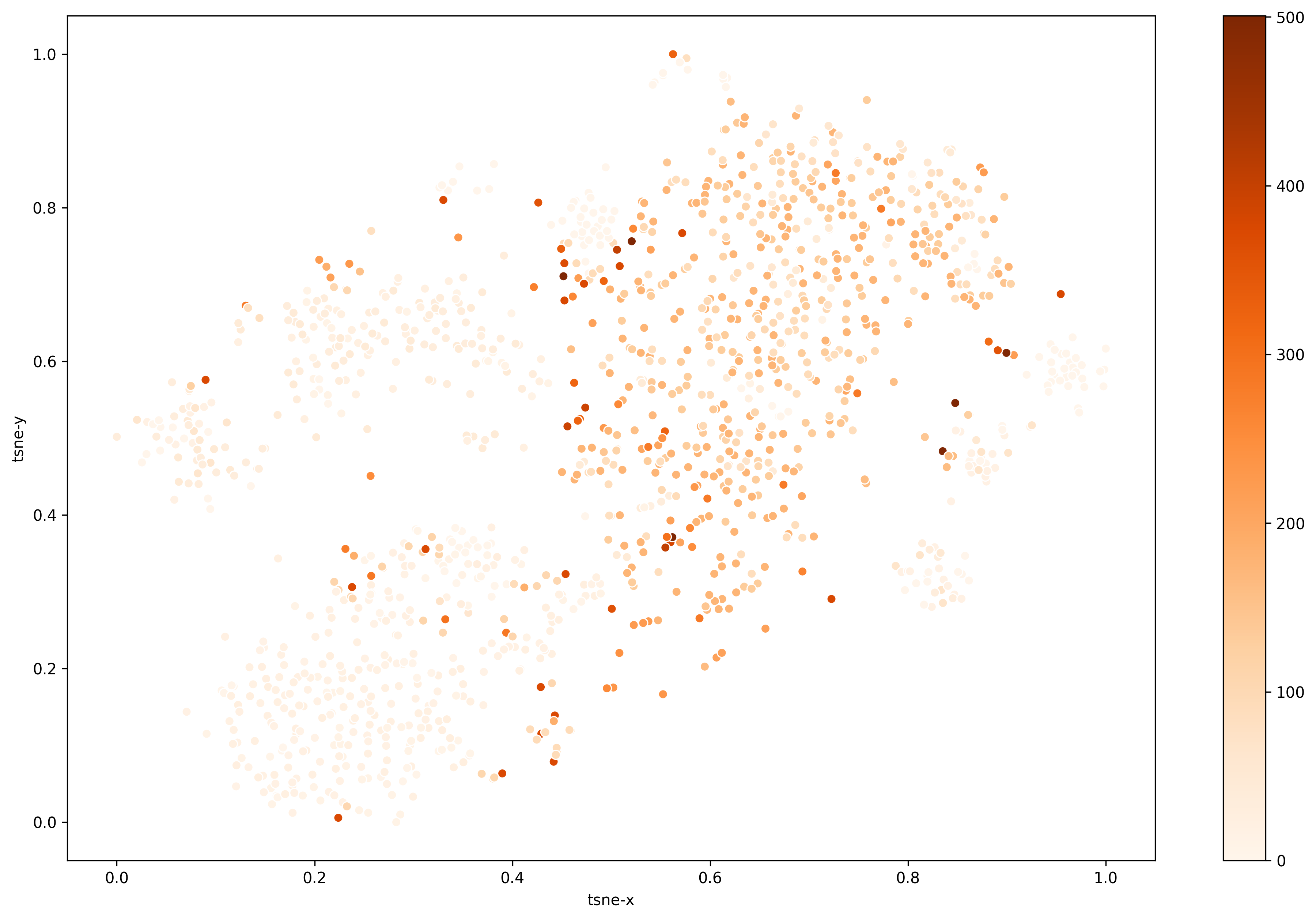} 
  \caption{CW} 
  \end{subfigure}
  \caption{t-SNE plots of the images generated by \phylonn{} and other baselines.}
  \label{fig:t-SNE}
 \vspace{-4ex} 
\end{figure}

\cref{fig:t-SNE} shows these constructed t-SNE plots with two different color-coding schemes. The first one (left column) color-codes the data-points based on the grouping of species at the second phylogenetic level (i.e., the direct ancestor of the specimen's species). Using this color-coding scheme allows us to inspect how different species cluster in the embedding space. The second color-coding (right column) is the average phylogenetic distance between the data-point and its $k$-nearest neighbors (KNN), where $k=5$ in this setup. The higher the average distance (i.e., the darker the data-point's color), the more distant the specimen is from those $k$ specimen's that are closest to it in the quantized embedding space. This color-coding helps us spot how well the different species are separated from each other in the embedding space, which generally characterizes the quality of the encoding and its propensity for downstream tasks, such as classification.

From \cref{fig:t-SNE}, we can see that \phylonn{} (top row)  clusters the generated images better than vanilla VQGAN and CW as evident from its hierarchical clustering where the specimens belonging to the same species clump into small clusters and these clusters in turn clump into larger clusters (representing ancestor nodes) that have a singular color. This demonstrates that \phylonn{} is able to learn a phylogenetically-meaningful encoding, whereas the other base models' clustering is quite fuzzy and poorly characterizes any biological knowledge. Also, by looking at the right column, we can see that \phylonn{} commits very little clustering error in terms of its phylogenetic constraints because the average phylogenetic distance is low (almost zero) for the majority of points. This is in contrast to the other baselines where there is quite a high clustering error as seen from the ``heat'' of its scatter plot.


\section{Conclusions and Future Work}
In this work, we presented a novel approach of \phylonn{} for discovering biological traits related to evolution automatically from images in an unsupervised  manner without requiring any trait labels. The key novelty of our approach is to leverage the biological knowledge of phylogeny to structure the quantized embedding space of \phylonn{}, where different parts of the embedding capture phylogenetic information at different ancestry levels of the phylogeny. This enables our method to perform a variety of tasks in a biologically meaningful way such as species-to-species image translation and identifying the ancestral lineage of newly discovered unseen species. 

In the future, our work can be extended to include a larger number of embedding dimensions to improve the visual quality of generated images and can be applied to other image datasets beyond the fish dataset. Future work can  explore extensions of \phylonn{} to generate images of ancestor species or to predict images of species that are yet to be evolved. Future work can also focus on making the discovered Imageome sequences more explainable by understanding the correspondence of each quantized code with a region in the image space. Our work opens a novel area of research in grounding image representations using tree-based knowledge, which can lead to new research paradigms in other fields of science where images are abundant but labels are scarce.

\begin{acks}
This work was supported, in part, by NSF awards for the HDR Imageomics Institute (Award \# 2118240) and the Biology-guided Neural Network (BGNN) projects (Award \# 1940247, \# 1940322, \# 1940233, \# 2022042, \# 1939505). Access to computing facilities was provided by the Advanced Research Computing (ARC) Center at Virginia Tech.
\end{acks}

\bibliographystyle{ACM-Reference-Format}
\balance
\bibliography{kdd-paper}

\clearpage
\appendix

\section{Dataset} \label{app:data}
As mentioned in \cref{sec:data}, the images we use in this work come from a 38-species semi-balanced subset of a larger collection that participated in the Great Lakes Invasives Network Project \href{https://greatlakesinvasives.org/portal/index.php}{(GLIN)}. After further splitting this subset into training and test sets, we apply some pre-processing that is necessary for the neural network to yield best results. This pre-processing includes padding the images with the ImageNet mean RGB color. We also use data augmentation when training the base VQGAN model. This includes random horizontal flips, spatial shifts and rotations, and brightness and contrast changes.

\section{Phylogeny preprocessing} \label{app:phylo}

As mentioned in \cref{sec:phylogeny}, we use a phylogenetic tree to characterize the evolutionary distances between the species in our dataset. The phylogeny corresponding to the dataset described in \cref{sec:data} was obtained using \textit{opentree} (\url{https://opentree.readthedocs.io/en/latest/}) python package. Phylogeny processing and manipulation were done using \textit{ete3} (\url{http://etetoolkit.org/}) python package.

\begin{figure}[t]
  \centering
  \includegraphics[width=1.00\linewidth]{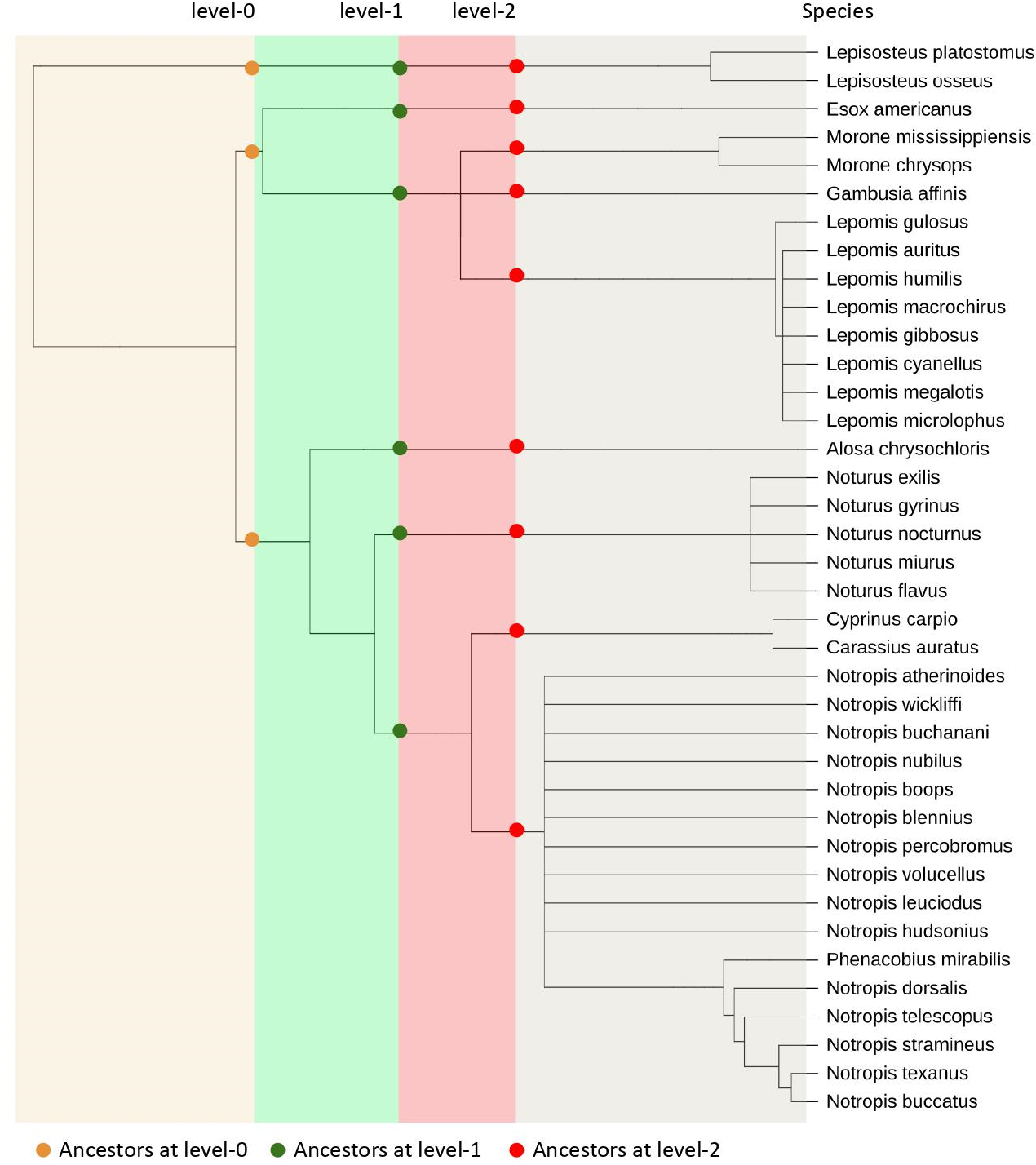}
  \caption{Ancestors at different levels of phylogeny}
  \label{fig:phylo-viz}
\end{figure}

In our application, we quantize our 38-species tree into $\nlevelsphylo{}=4$ distinct phylogenetic levels. Each level groups the 38 species based on their common ancestry within that level. 
\cref{fig:phylo-viz} outlines each level with its corresponding species groupings.


\section{Hyper-parameter selection}
In terms of hyper-parameter tuning, we used the following settings for each of the trained models:

\noindent \textbf{Vanilla VQGAN:} We trained a VQGAN with a codebook of 1024 possible codes and an embedding sequence of 256 codes. We trained the model for $836$ epochs with a learning rate of $4.5 \times 10^{-6}$. We used this VQGAN as the base model for the rest of the models. A batch size of 32 was used.

\noindent \textbf{\phylonn{}} Taking the base VQGAN model described above, we trained a \phylonn{} that has $\zphyloq$ of dimensions $(\nlevelsphylo= 4,\codesperlevel = 8)$. $\znonattrq$ also has the same dimensionality. The dimensionality of the embedding itself is $\embeddingsize = 16$ , and the size of the codebook $\ncodebook = 64$. A batch size of 32 was used.

\noindent \textbf{Concept Whitening (CW)} Taking the base VQGAN model described above, we trained CW for $20$ epochs using the same hyper-parameters as Vanilla VQGAN. We used a batch size of 20 for the concepts.

\noindent \textbf{Latent Space Factorization (LSF):} With the base VQGAN model described above, a variational autoencoder was introduced between the base encoder and the quantization layer. The model was trained for 200 epochs with a learning rate of $1 \times 10^{-4}$. We used an embedding dimension of 1024 for the variational autoencoder. 


\section{Details of Morphological Distance Processing}
\label{app:morph_dist}
The 8 functionally relevant traits that we used from the FishShapes dataset include: standard length, maximum body depth, maximum fish width, lower jaw length, mouth width, head depth, minimum caudal peduncle depth, and minimum caudal peduncle. Some species were not available in the FishShapes dataset, so when possible, the closest relative was substituted. (\textit{Notropis percobromus} was replaced with \textit{Notropis rubellus}, and \textit{Carassius auratus} was replaced with \textit{Carassius carassius}). Also, two species of \textit{Lepisosteus} had no close relatives and were thus removed fro the Spearman correlation analysis.  To correct for overall size and allometry, each measurement was log transformed and regressed against Standard Length (SL) using a phylogenetic regression in the R package \textit{phylolm}, with the residuals from the regression being the inputs into PACA. Distances in the principal components of PACA were measured as the Mahalonobis distance between the multivariate means using a covariance matrix proportional to the evolutionary rate matrix from the multivariate Brownian Motion fit in the R package mvMORPH \cite{clavel2015}. 

\section{t-SNE plots for test images}

\begin{figure}[h]

   \centering
     \begin{subfigure}[b]{1.0\columnwidth}
  \centering
  \includegraphics[width=0.47\textwidth]{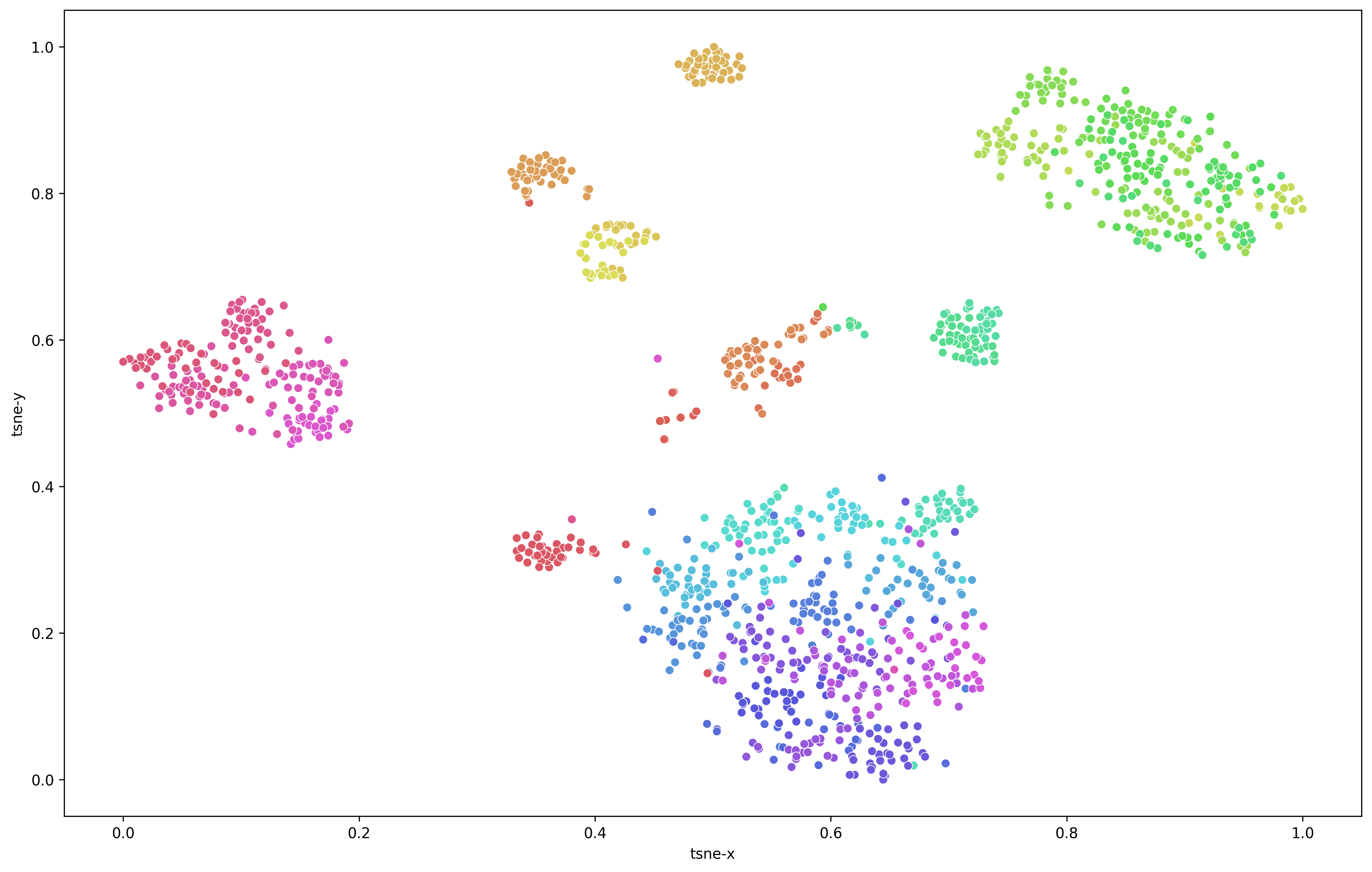}
  \hfill
  \includegraphics[width=0.47\textwidth]{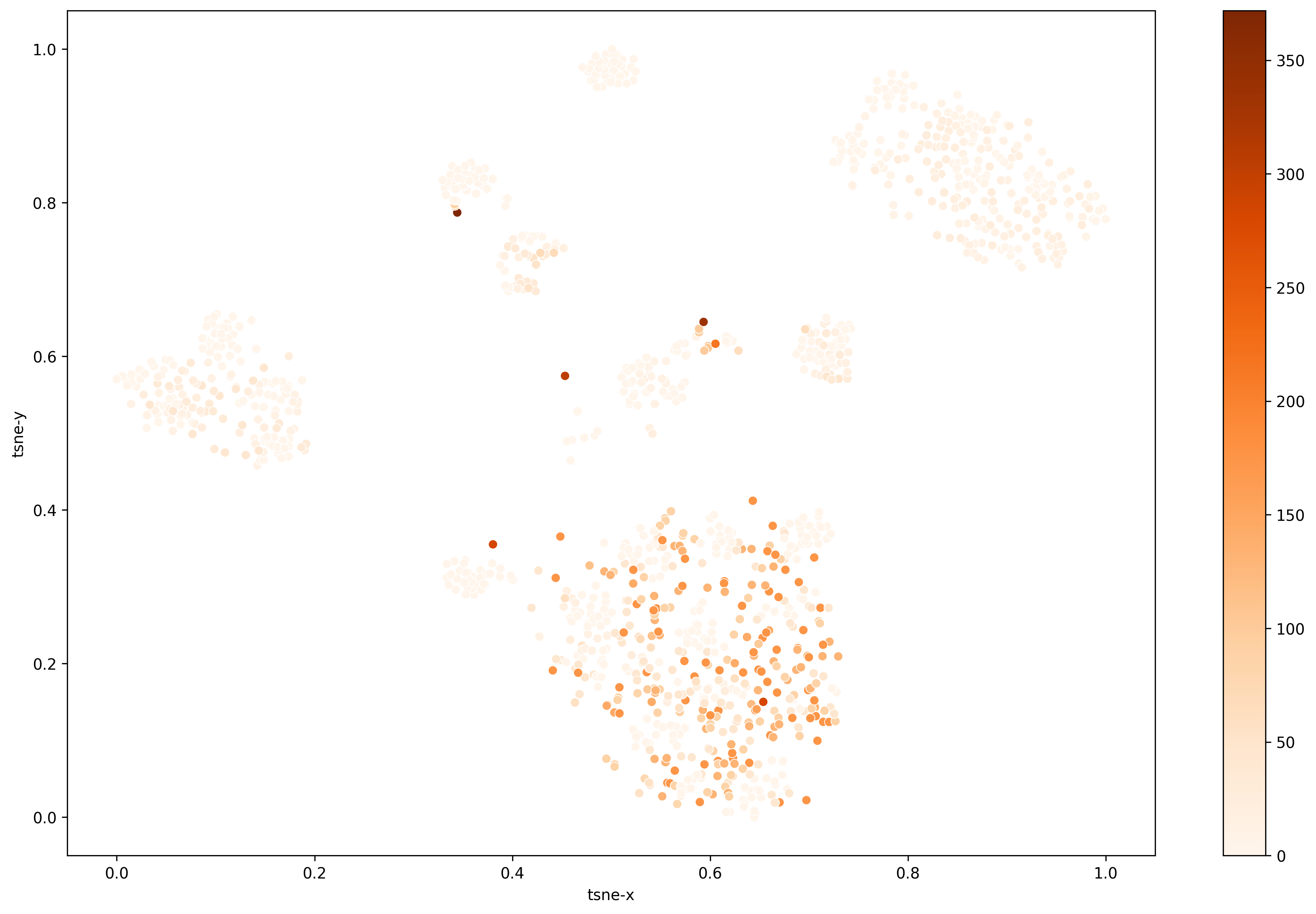}
  \caption{\phylonn{}} 
  \end{subfigure} 
  \vskip\baselineskip
  \begin{subfigure}{1.0\columnwidth} 
  \includegraphics[width=0.47\textwidth]{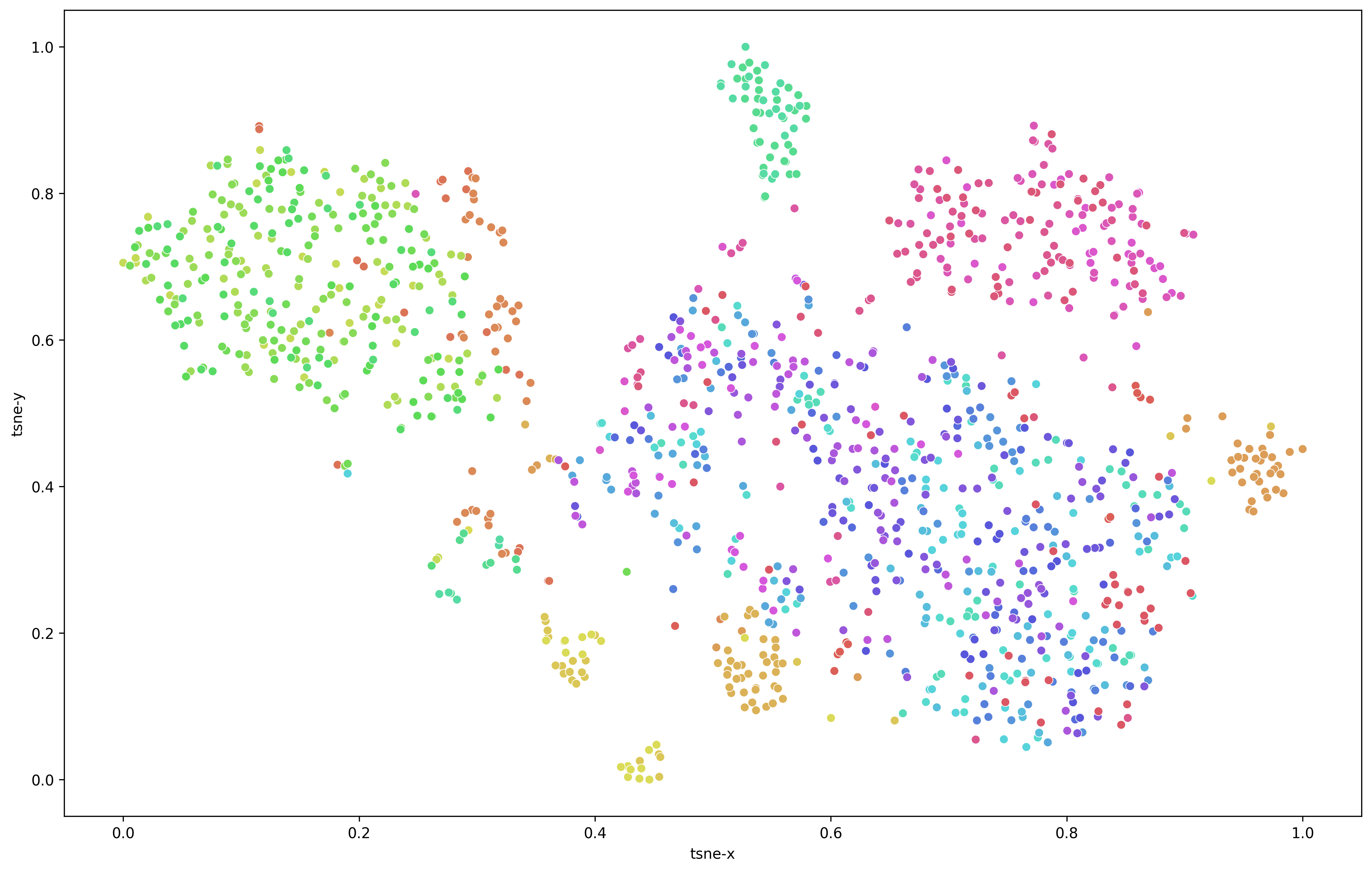}
  \hfill
  \includegraphics[width=0.47\textwidth]{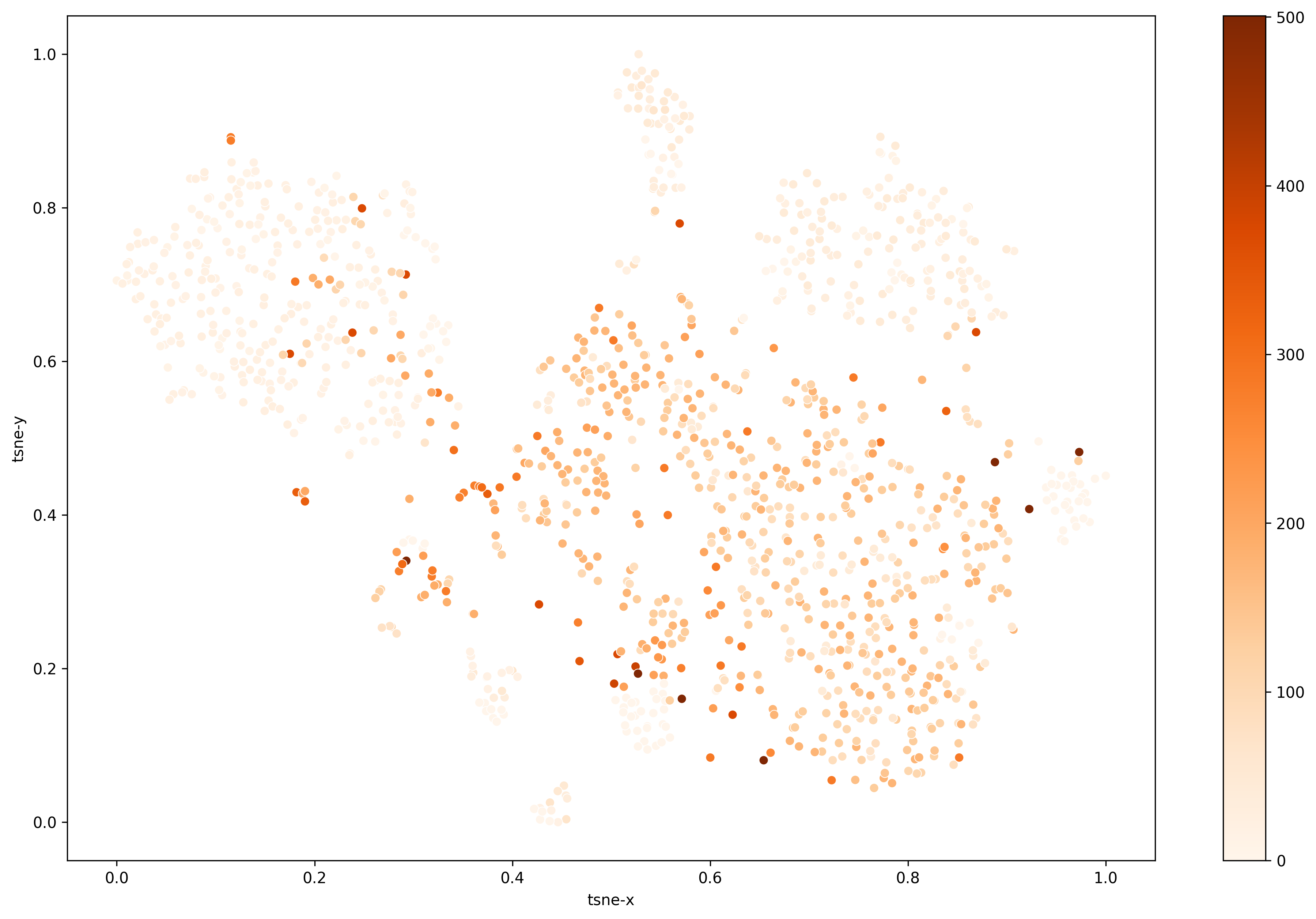} 
  \caption{Vanilla VQGAN} 
  \end{subfigure}
\vskip\baselineskip
  \begin{subfigure}[b]{1.0\columnwidth}
  \includegraphics[width=0.47\textwidth]{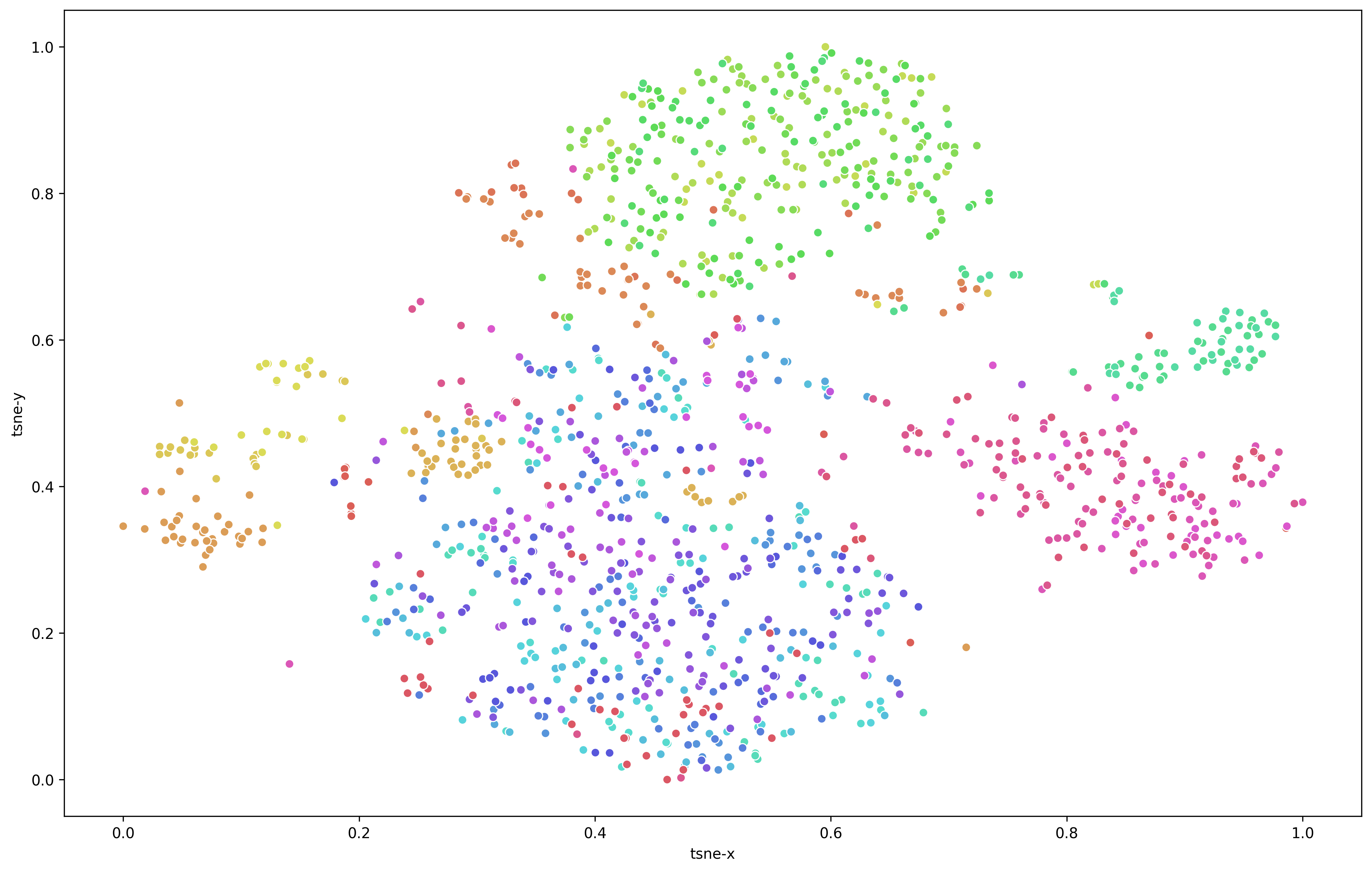}
  \hfill
  \includegraphics[width=0.47\textwidth]{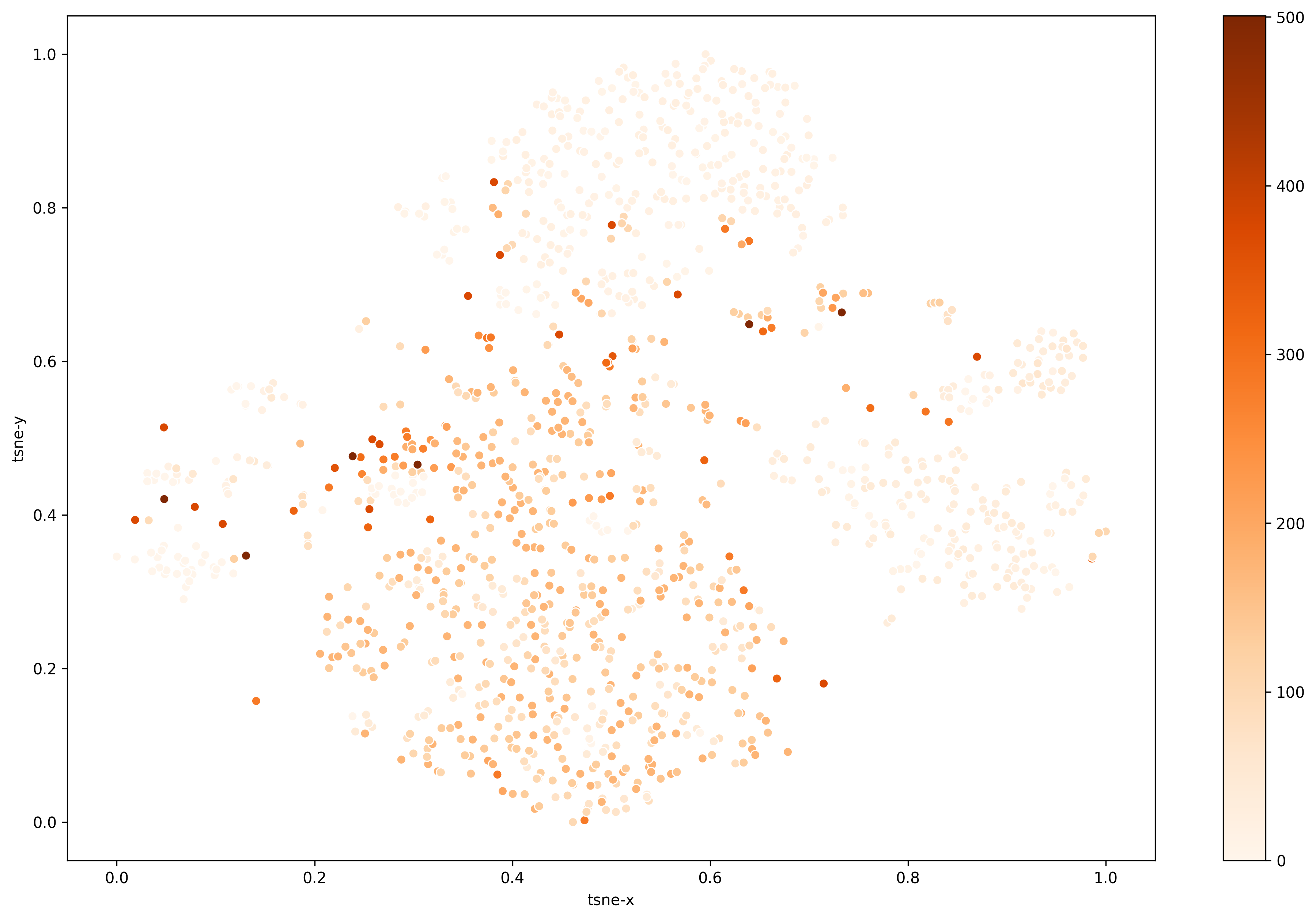}
  \caption{CW} 
  \end{subfigure} 
  \caption{t-SNE plots of the test set images using different models}
      \label{fig:t-SNE-test-images}
\end{figure}

As we have shown in \cref{ss:tsne}, the embedding of the images generated by our \phylonn{} model are more meaningful than those generated by a vanilla VQGAN. In this section, we run the same analysis on the test images. Clearly, a similar case can be made here, namely that the encoding of the images using \phylonn{} is superior to other models' in terms of its clustering. \cref{fig:t-SNE-test-images} illustrates this point clearly when comparing \phylonn{}'s (top row) t-SNE plots with those of the other models'. Both vanilla VQGAN and CW perform worse at clustering the dataset compared to \phylonn{}. 

\section{Examples of Phylo Histograms} \label{app:hist}

In \cref{ss:unseen}, by means of calculating the average JS-divergence of sequence histograms, we investigated how well the Imageome sequences match for species that share a common ancestor, as opposed to those that don't. In this section, we show some of these histogram plots to illustrate their value and the insight they could provide. 

\begin{figure}[h]

   \centering
        \begin{subfigure}[b]{1.0\columnwidth}
    \includegraphics[width=\textwidth]{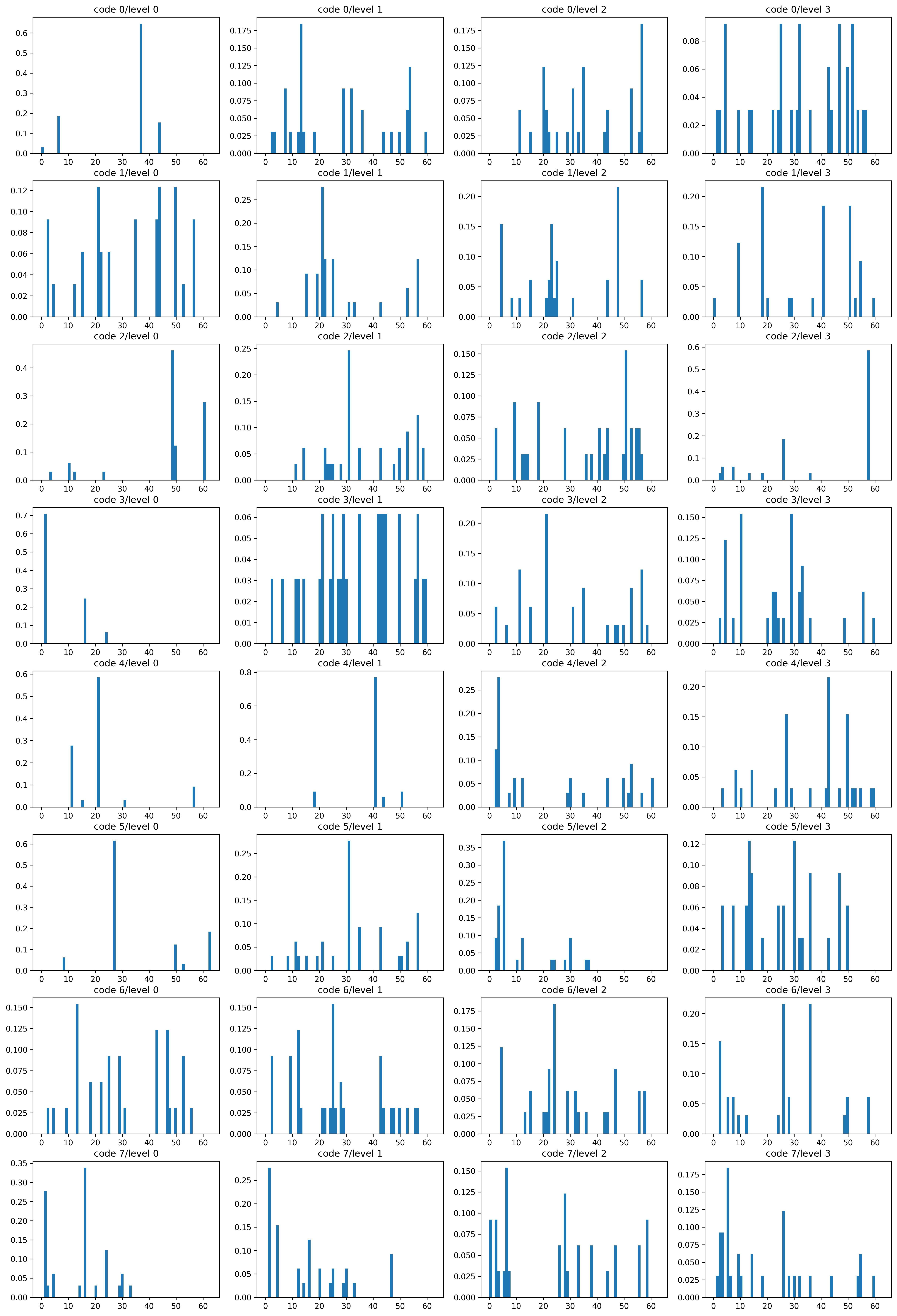}
  \end{subfigure} 
    \caption{\textit{\textit{Notropis nubilus}} }
  \label{fig:histograms1}
\end{figure}

\begin{figure}[h]

   \centering
        \begin{subfigure}[b]{1.0\columnwidth}
    \includegraphics[width=\textwidth]{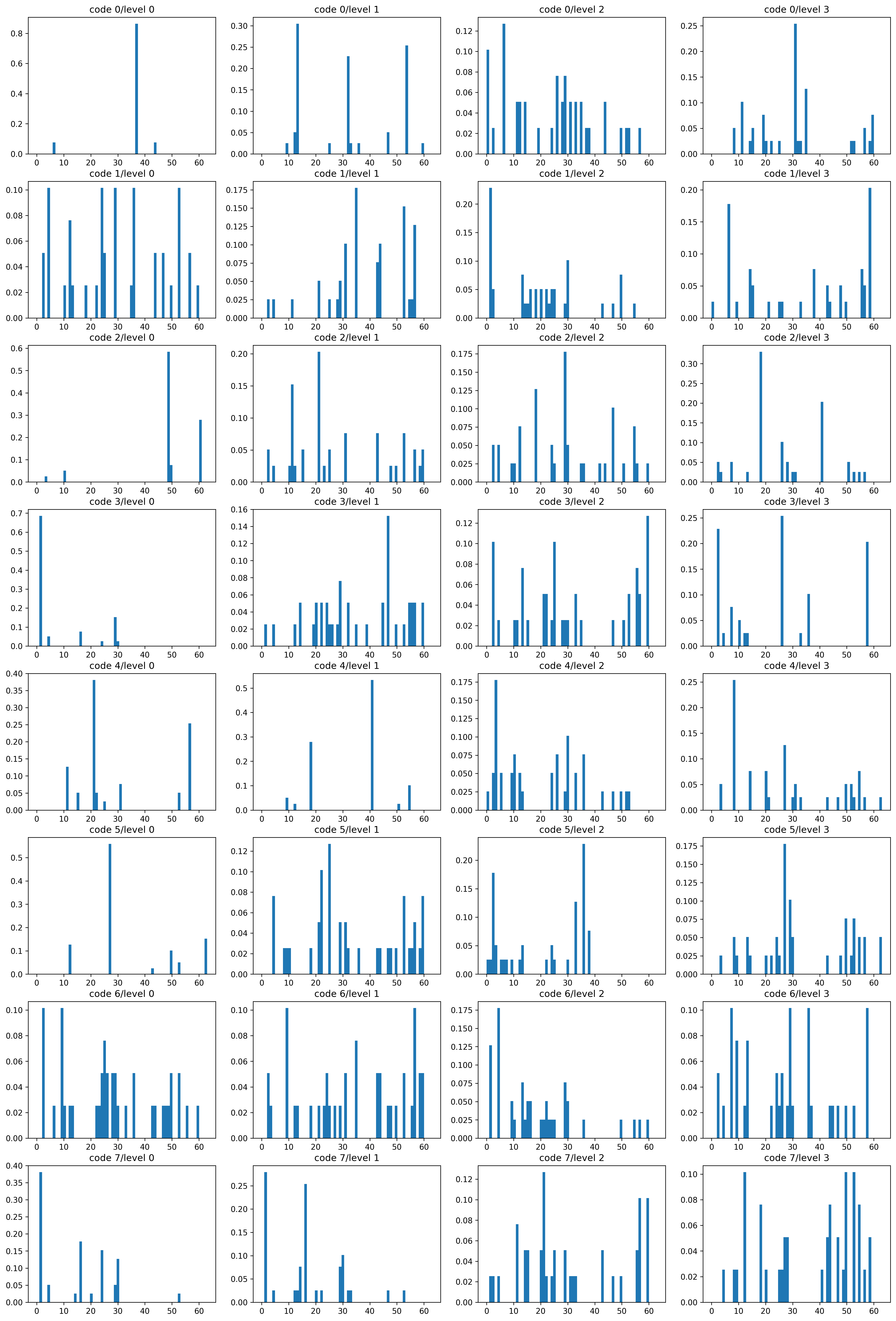}
  \end{subfigure} 
      \caption{\textit{\textit{Notropis percobromus}} }
  \label{fig:histograms2}
\end{figure}

\begin{figure}[h]

   \centering
        \begin{subfigure}[b]{1.0\columnwidth}
    \includegraphics[width=\textwidth]{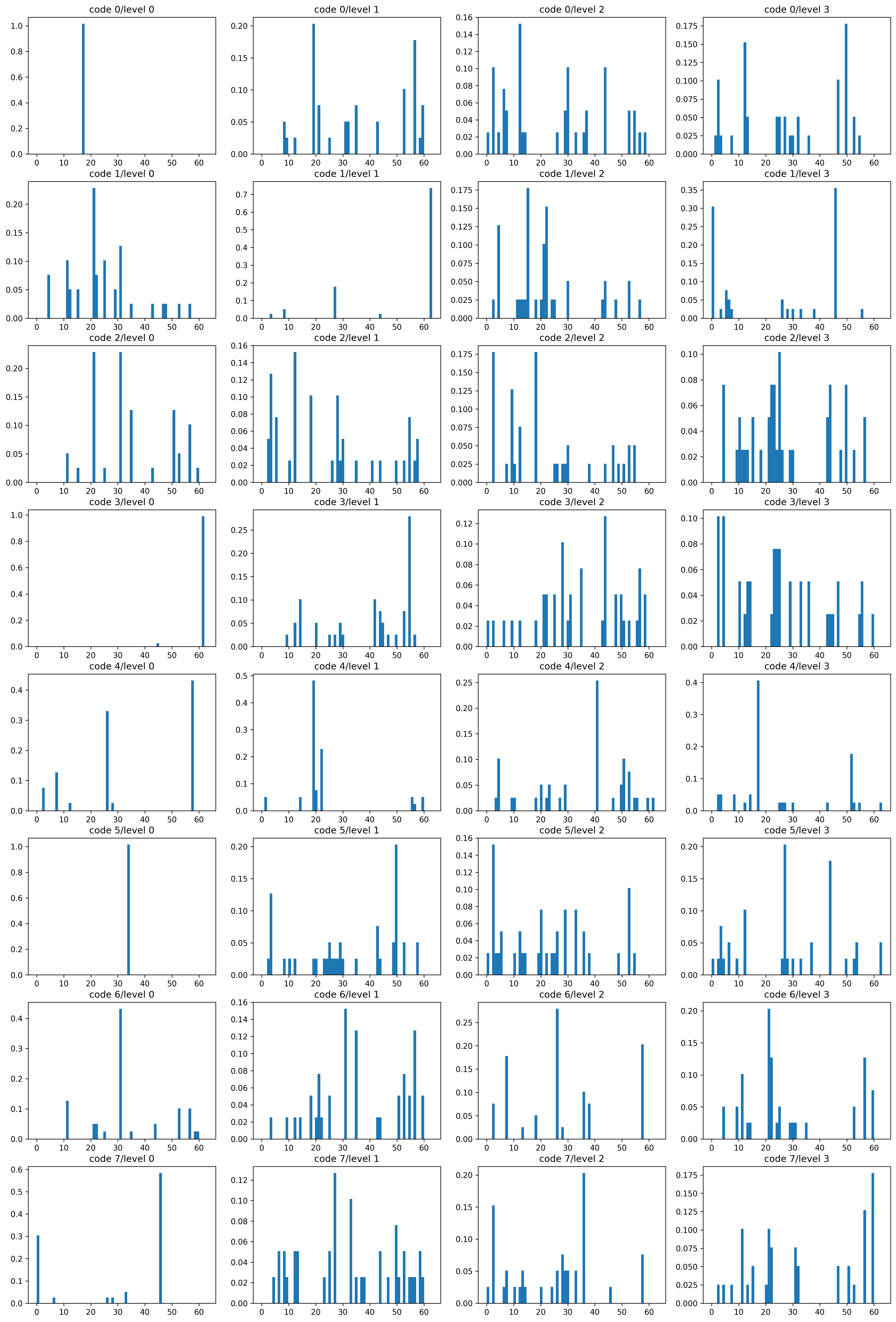}
  \end{subfigure} 
    \caption{\textit{\textit{Lepomis macrochirus}} }
  \label{fig:histograms3}
\end{figure}

In \cref{fig:histograms1,fig:histograms2,fig:histograms3}, each histogram represents a code location in the phylogenetic sequence. Each column represents one of the $\nlevelsphylo=4$ phylogenetic levels into which the phylogeny was quantized, starting with the species level from right and climbing the phylogeny all the way to the earliest ancestral level on the left. Each column has $\codesperlevel=8$ codes. Each histogram shows the relative frequency of each code of the learned $\ncodebook = 64$ codes for its corresponding sequence location. The lower a histogram's entropy (i.e., when there is only one or a couple of codes that dominate the histogram's frequency spectrum), the more important that code location hypothetically is for characterizing the species at its corresponding phylogenetic level.

As can be seen, both \textit{Notropis} species share many codes at many sequence locations up to, but not including, the species level. This is because these species share an immediate ancestor. In contrast, We can see that the \textit{Lepomis} species has a distinct histogram signature and does share almost no codes with the \textit{Notropis} species, except for the earlier ancestral level (i.e., the left column).

\clearpage

\newpage

\end{document}